\documentclass[twocolumn]{article} 

\usepackage[letterpaper, margin=0.75in, columnsep=0.24in]{geometry}

\usepackage{hyperref}
\usepackage{graphicx}
\usepackage{amsmath}
\usepackage{amssymb}
\usepackage{cite}
\usepackage{url}
\usepackage{multirow}
\usepackage{stmaryrd}
\usepackage{makecell}
\usepackage{algorithm}
\usepackage{algorithmic}

\usepackage{booktabs}

\usepackage{sectsty}

\sectionfont{\large}
\subsectionfont{\large}
\subsubsectionfont{\normalsize}

\usepackage[affil-it]{authblk}

\global\long\def\bx{\mathbf{\mathbf{x}}}%

\global\long\def\bx{\mathbf{x}}%
\global\long\def\and{\cap}%

\begin{document}

\title{CapsFake: A Multimodal Capsule Network\\for Detecting Instruction-Guided Deepfakes}

\author{Tuan Nguyen}
\author{Naseem Khan}
\author{Issa Khalil}

\affil{Qatar Computing Research Institute, Hamad Bin Khalifa University}

\date{}

\maketitle

\begin{abstract}
The rapid evolution of deepfake technology, particularly in instruction-guided image editing, threatens the integrity of digital images by enabling subtle, context-aware manipulations. Generated conditionally from real images and textual prompts, these edits are often imperceptible to both humans and existing detection systems, revealing significant limitations in current defenses. We propose a novel multimodal capsule network, CapsFake, designed to detect such deepfake image edits by integrating low-level capsules from visual, textual, and frequency-domain modalities. High-level capsules, predicted through a competitive routing mechanism, dynamically aggregate local features to identify manipulated regions with precision. Evaluated on diverse datasets, including MagicBrush, Unsplash Edits, Open Images Edits, and Multi-turn Edits, CapsFake outperforms state-of-the-art methods by up to 20\% in detection accuracy. Ablation studies validate its robustness, achieving detection rates above 94\% under natural perturbations and 96\% against adversarial attacks, with excellent generalization to unseen editing scenarios. This approach establishes a powerful framework for countering sophisticated image manipulations.

\end{abstract}


\section{Introduction}
\begin{figure*}
\begingroup
\setlength{\tabcolsep}{2pt} 
\renewcommand{\arraystretch}{1.0} 
\centering
\begin{tabular}{cccc}
\includegraphics[width=0.2\textwidth]{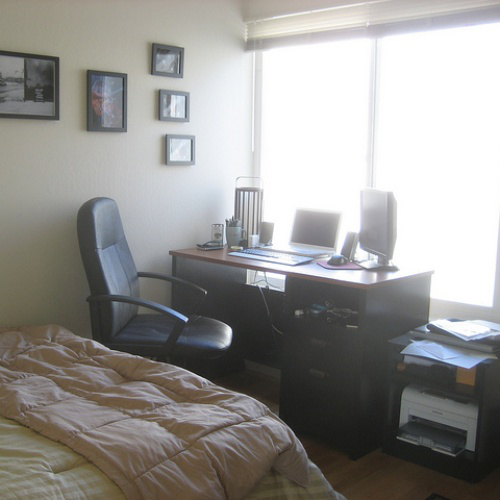} &
\includegraphics[width=0.2\textwidth]{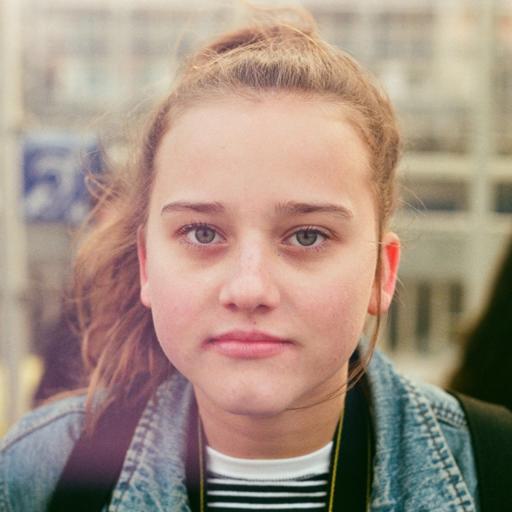} & 
\includegraphics[width=0.2\textwidth]{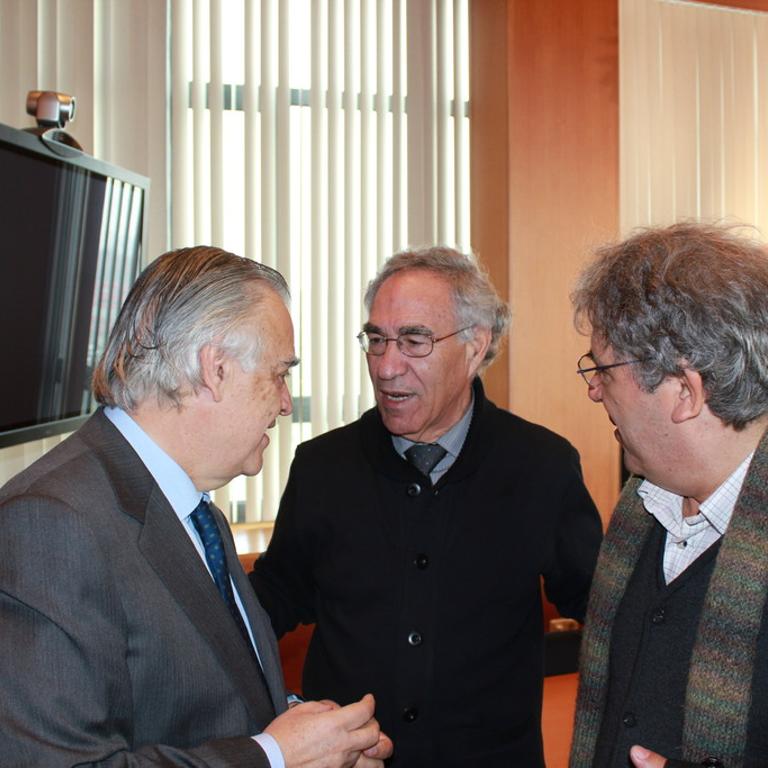} &
\includegraphics[width=0.2\textwidth]{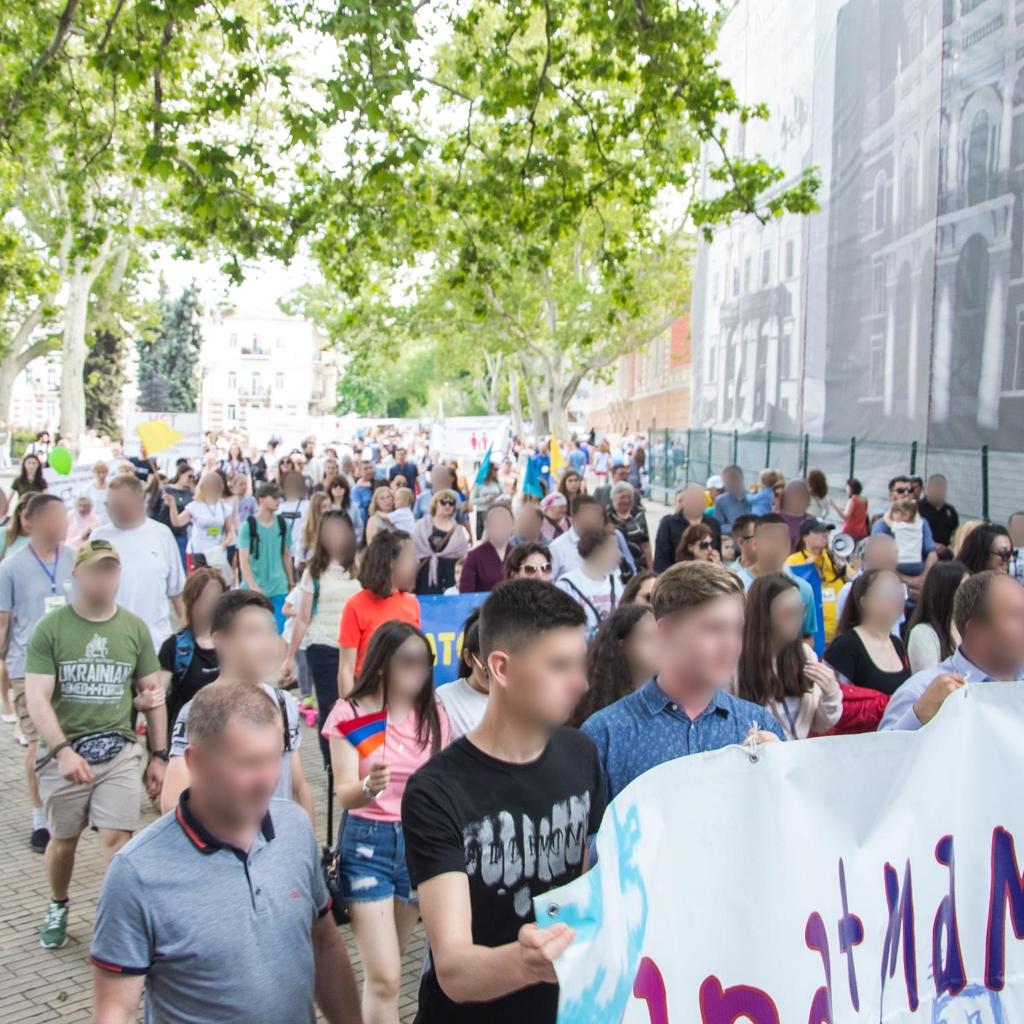} \tabularnewline

\begin{tabular}{c}
\includegraphics[width=0.2\textwidth]{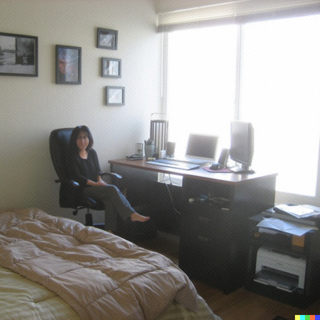} \\
\scriptsize (a) MagicBrush
\end{tabular} &
\begin{tabular}{c}
\includegraphics[width=0.2\textwidth]{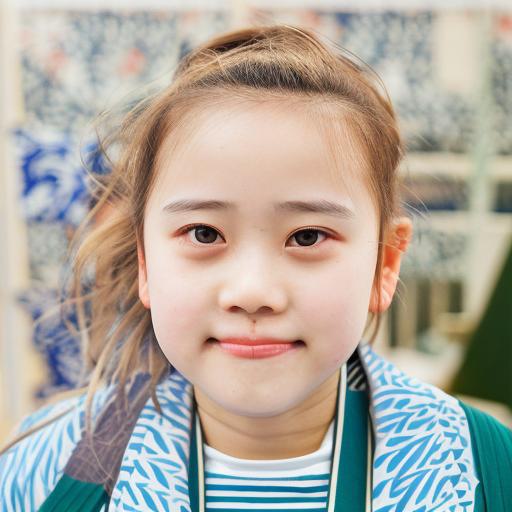} \\
\scriptsize (b) Unsplash Edits
\end{tabular} &
\begin{tabular}{c}
\includegraphics[width=0.2\textwidth]{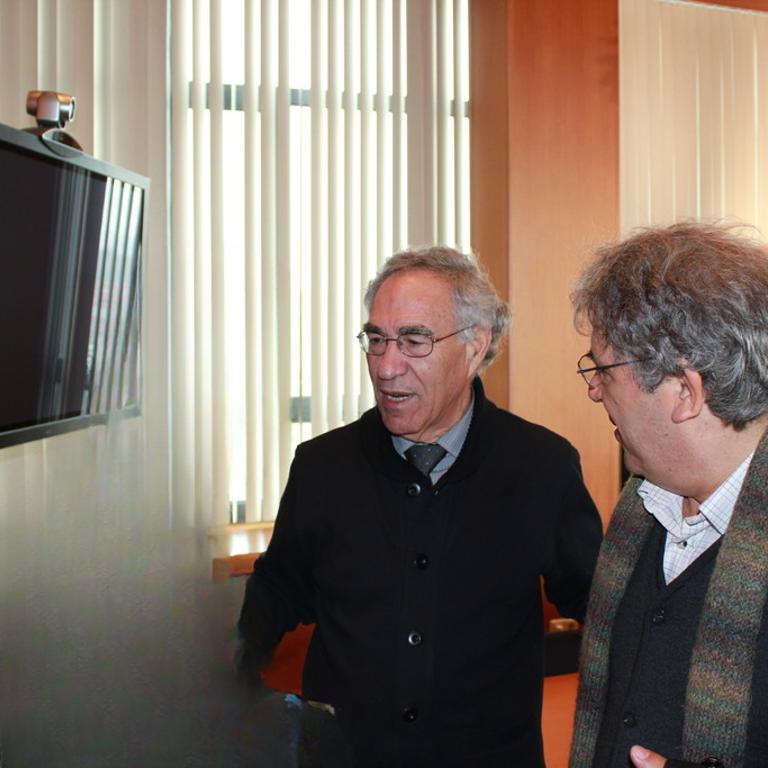} \\
\scriptsize (c) Open Images Edits
\end{tabular} &
\begin{tabular}{c}
\includegraphics[width=0.2\textwidth]{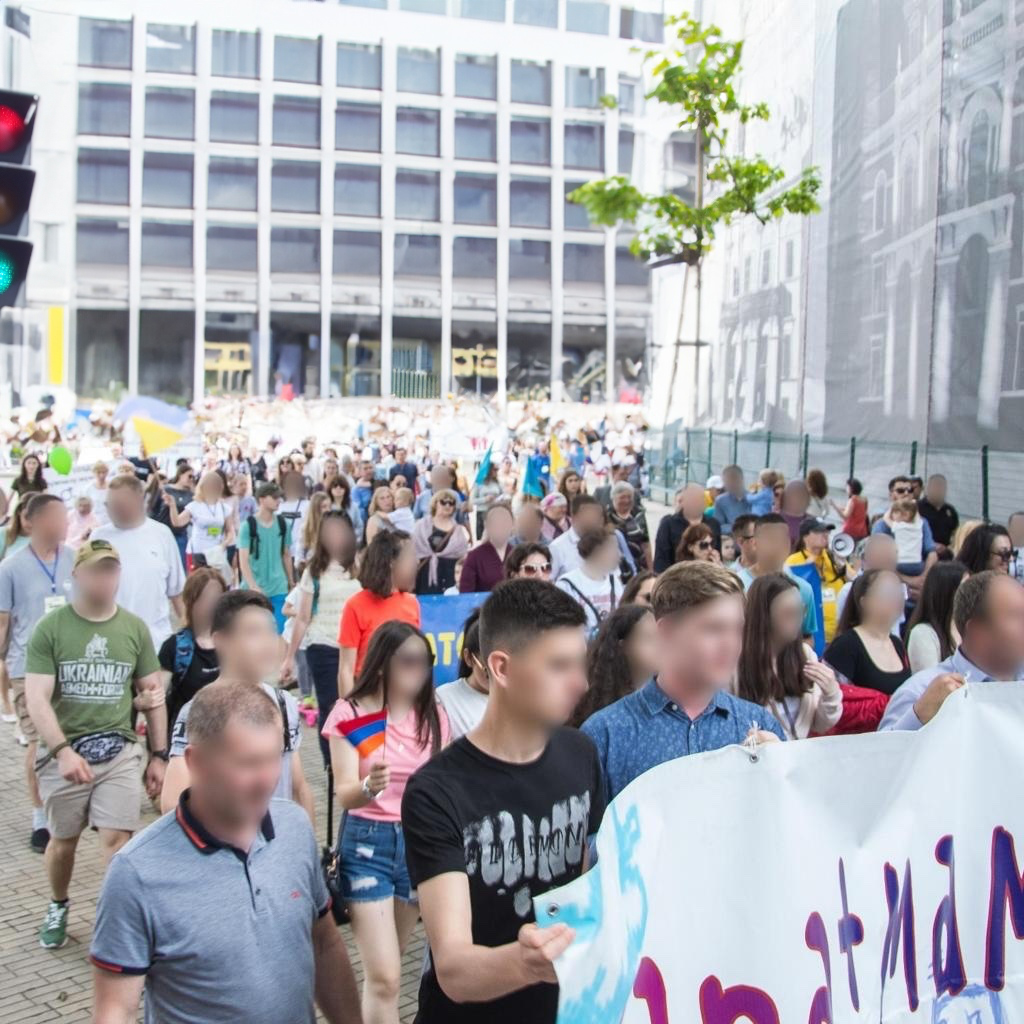} \\
\scriptsize (d) Multi-turn Edits
\end{tabular}
\end{tabular}
\caption{Real (first row) and fake (second row) image editing pairs, sourced from the Unsplash Edits (a), the Open Images Edits (b), MagicBrush (c), and MagicBrush (d).}
\label{fig:examples_image_edit}
\endgroup
\end{figure*}

The rapid advancement of deepfake technology poses a significant threat to the authenticity of digital images, with profound implications for democratic institutions, national security, and social trust. Increasingly sophisticated manipulation techniques can now fabricate evidence, impersonate public figures, and disseminate misinformation at unprecedented scale and speed. Instruction-guided image editing—a particularly insidious form of deepfake generation—enables subtle, context-driven alterations to real images via simple textual prompts. Unlike fully synthetic images, these edits maintain high visual coherence with the original by selectively changing specific elements—such as modifying an object's color, removing identifiable features, or inserting new objects—while preserving the surrounding context. As illustrated in Figure \ref{fig:examples_image_edit}, these manipulations often remain imperceptible to human observers while fundamentally altering the semantic meaning of visual content, creating a critical vulnerability in our information ecosystem that demands robust technological countermeasures.


\begin{table}
\begin{centering}
\caption{Deepfake detection performance (\%) on Open Images Edits.}\label{tab:baseline-performance}
\resizebox{\columnwidth}{!}{%
    \begin{tabular}{ccccc}
    \hline
    Methods & Precision & Recall & F1 & Accuracy\tabularnewline
    \hline
    UniCLIP \cite{ojhaUniversalFakeImage2024} & 71.75 & 53.93 & 61.58 & 66.35\tabularnewline
    DCT \cite{rickerDetectionDiffusionModel2024} & 77.04 & 41.68 & 54.09 & 64.63\tabularnewline
    ZeroFake \cite{shaZeroFakeZeroShotDetection} & 57.53 & 27.11 & 36.85 & 53.55\tabularnewline
    DE-FAKE \cite{shaFAKEDetectionAttribution2023} & 72.03 & 46.32 & 56.38 & 64.17\tabularnewline
    ObjectFormer \cite{wang2022objectformer} & 90.20 & 58.32 & 67.33 & 71.71\tabularnewline
    UnivConv2B \cite{abdullahAnalysisRecentAdvances2024} & 63.58 & 77.65 & 69.91 & 66.58\tabularnewline
    \hline
    \end{tabular}
}
\par\end{centering}
\end{table}

Recent state-of-the-art (SOTA) deepfake detectors, including UnivCLIP \cite{ojhaUniversalFakeImage2024}, DCT \cite{rickerDetectionDiffusionModel2024}, ZeroFake \cite{shaZeroFakeZeroShotDetection}, DE-FAKE \cite{shaFAKEDetectionAttribution2023} and UnivConv2B \cite{abdullahAnalysisRecentAdvances2024}, have shown remarkable success in identifying fully synthetic images generated by models like Stable Diffusion \cite{rombachHighResolutionImageSynthesis2022}, DALL-E \cite{rameshHierarchicalTextConditionalImage2022}, and StyleCLIP \cite{patashnikStyleCLIPTextDrivenManipulation2021a}. However, their efficacy dramatically diminishes when confronted with instructional image edits, where manipulations are precisely targeted and visually coherent with original content. As demonstrated in Table \ref{tab:baseline-performance}, experiments on the Open Images Edits dataset \cite{geSEEDDataEditTechnicalReport2024} reveal that even state-of-the-art methods achieve critically low recall—DE-FAKE (46.32\%), ZeroFake (27.11\%), and UnivConv2B (77.65\%)—despite their strong performance on fully synthetic content. This substantial security gap stems from two fundamental limitations: insufficient cross-modal integration that fails to capture semantic inconsistencies across complementary information channels, and architectural constraints \cite{liuConvNet2020s2022} that prevent effective preservation of spatial relationships between authentic and manipulated content regions.

Current multi-modal frameworks like DE-FAKE \cite{shaFAKEDetectionAttribution2023} and ObjectFormer \cite{wang2022objectformer} employ feature concatenation or attention mechanisms that fail to effectively model the complex relationships between modalities, particularly when manipulations selectively target specific semantic elements while preserving others. Simultaneously, the convolutional architectures \cite{krizhevskyImageNetClassificationDeep2012, ronnebergerUNetConvolutionalNetworks2015, heDeepResidualLearning2015, renFasterRCNNRealTime2016} underlying these approaches rely on max-pooling operations that discard critical spatial information and feature relationships, making them fundamentally ill-suited for identifying subtle manipulations that maintain global context while altering security-critical details. Capsule networks \cite{sabourDynamicRoutingCapsules2017} directly address both vulnerabilities by: (1) encoding entities as vectors that capture characteristic properties and existence probabilities, and (2) implementing dynamic routing that selectively aggregates information across modalities based on agreement among low-level features. This dual capability enables effective detection of localized manipulations by preserving both cross-modal semantic relationships and spatial hierarchies essential for distinguishing between authentic variations and malicious edits.

We propose \textbf{CapsFake}, a multimodal capsule network that implements a semantic integrity verification framework for detecting instruction-guided image manipulations. CapsFake performs dynamic cross-modal routing to enable entity-level reasoning about semantic consistency between visual elements, textual descriptions, and frequency signatures. Through this architecture, low-level capsules encoding domain-specific properties compete to predict high-level semantic entities, creating an intrinsic verification system that identifies manipulations by detecting modal disagreement at precise spatial locations. This approach directly addresses the fundamental security challenge of identifying semantically significant edits that preserve visual coherence with authentic content—a capability beyond the reach of current detection paradigms.

Our contributions are as follows:
\begin{itemize}
    \item \textbf{Threat Pipeline Formulation.} We formalize a novel threat model for \textit{instruction-guided malicious image manipulation}, outlining how generative AI can be leveraged to produce context-aware, semantically meaningful image edits that preserve visual coherence while enabling identity forgery, evidence tampering, and misinformation.

    \item \textbf{Capsule-Based Security Architecture.} We present the first application of capsule networks for deepfake detection in the context of instruction-guided manipulations. Our proposed framework, CapsFake, performs fine-grained reasoning over visual, textual, and frequency modalities using a multimodal capsule routing mechanism, enabling robust and interpretable detection of localized semantic edits. Code and benchmarks are available at: \url{https://github.com/tuanrpt/CapsFake.git}.

    \item \textbf{Superior Detection Across Diverse Threat Scenarios.} CapsFake consistently outperforms state-of-the-art detectors on challenging benchmarks, including MagicBrush, Unsplash Edits, Open Images Edits, and Multi-turn Edits. It achieves up to 99.5\% F1 and 100\% recall on visually coherent edits, and maintains leading performance across increasingly complex multi-round manipulations.

    \item \textbf{Comprehensive Security Evaluation.} We evaluate CapsFake under natural perturbations and white-box adversarial attacks (FGSM and PGD), achieving over 94\% F1 in all natural degradation settings and up to 100\% recall under adversarial attack. Additionally, we demonstrate strong robustness against black-box transfer attacks and semantic adversaries, confirming the model’s cross-modal consistency and real-world deployability.
\end{itemize}

This security-centered approach bridges the critical gap in digital media authentication, providing a robust defense against emerging threats from increasingly accessible and sophisticated instruction-guided manipulation tools.

\section{Threat Model and Security Objectives}
To provide a clearer understanding of how CapsFake addresses modern security challenges in deepfake detection, we outline our threat model along with the detector's core goals and capabilities. These design objectives reflect practical requirements for secure deployment in real-world settings, where robustness, generalization, and interpretability are essential for trustworthy decision-making systems.
\paragraph{Achieving High Detection Performance on Semantic Deepfake Manipulations.}
The primary objective of \textbf{CapsFake} is to achieve robust and accurate detection of the most challenging form of deepfake content: \textit{instruction-guided deepfakes} (also referred to as \textit{semantic deepfakes}). These manipulations involve subtle, instruction-guided edits that preserve global visual coherence while altering critical semantic content, often escaping human perception.
\paragraph{Robustness to Natural Distortions and Adversarial Perturbations.}
In real-world scenarios, images may undergo transformations such as compression, noise, lighting changes, or minor distortions. CapsFake is designed to maintain detection performance under these conditions, exhibiting high robustness across a variety of natural perturbations. Additionally, we evaluate its adversarial resilience under both \textit{white-box} and \textit{black-box} attack settings. In the white-box setting, we assume the attacker has full access to the model's architecture and parameters, and employs adversarial techniques such as Fast Gradient Sign Method (FGSM)~\cite{goodfellowExplainingHarnessingAdversarial2015} and Projected Gradient Descent (PGD)~\cite{madryDeepLearningModels2019}. In the black-box setting, the attacker crafts adversarial examples using other known state-of-the-art detectors, aiming to transfer the attack to CapsFake. CapsFake should demonstrate strong robustness under both conditions.

\paragraph{Generalization to Unseen Domains.}
CapsFake is built to generalize beyond its training distribution. We test its ability to detect manipulations in unseen domains by training on one dataset and evaluating on others containing distinct editing styles and content distributions. This setting simulates practical deployment conditions, where manipulation techniques continuously evolve.

\paragraph{Localization of Semantic Manipulations.}
An advantage of CapsFake is its ability to localize manipulated regions within an image while maintaining a global understanding of content. This allows the detector not only to predict whether an image is fake, but also to highlight suspicious areas, offering interpretability and aiding downstream forensic analysis. By capturing localized inconsistencies, CapsFake provides actionable insights for secure image verification.

\section{Related Work}\label{sec:related_work}
This section analyzes existing deepfake detection methods through a security lens, highlighting their limitations against instruction-guided image manipulations and positioning our contribution within the evolving threat landscape.

\subsection{Unimodal Detection Approaches}
Early deepfake detection relied primarily on single-modality approaches, which exhibit significant security vulnerabilities when confronted with sophisticated manipulation techniques. Visual-only detectors such as UnivCLIP \cite{ojhaUniversalFakeImage2024} leverage pretrained vision-language model embeddings, demonstrating strong generalization across generative models but failing to detect subtle manipulations that preserve global visual coherence. When faced with instruction-guided edits that selectively modify objects while maintaining overall image context, these detectors exhibit substantially degraded performance.

Frequency-domain detectors \cite{rickerDetectionDiffusionModel2024} analyze transform-space artifacts characteristic of generative models, particularly diffusion-based approaches. While effective against full-image synthesis, these methods are vulnerable to targeted attack vectors that preserve frequency characteristics in manipulated regions. Their effectiveness diminishes significantly when confronted with localized, semantically meaningful edits that maintain frequency signatures consistent with authentic content \cite{abdullahAnalysisRecentAdvances2024}.

ZeroFake \cite{shaZeroFakeZeroShotDetection} proposes a distinct approach to deepfake detection by leveraging DDIM inversion and reconstruction processes to identify fake content without the need for training data. It compares reconstruction distances between original images and their adversarially-prompted counterparts to detect fully synthetic content. However, this approach faces critical limitations in security-sensitive contexts involving instruction-guided edits, where the subtle and localized nature of manipulations reduces the reconstruction gap, making detection significantly more difficult. 

Handcrafted feature-based approaches \cite{zhengSiameseMultilayerPerceptrons2015, baiGrowingRandomForest2017} exhibit even greater security limitations, as they are designed for specific artifact patterns that sophisticated editing techniques explicitly avoid. The rapid evolution of generative models continually invalidates these static detection strategies, creating a fundamental security vulnerability in deployed systems \cite{masoodDeepfakesGenerationDetection2021}.

\subsection{Multi-modal Detection Frameworks}
Recognizing the limitations of unimodal approaches, researchers have proposed multi-modal frameworks to enhance detection robustness. DE-FAKE \cite{shaFAKEDetectionAttribution2023} combines visual and textual embeddings to identify semantic inconsistencies between image content and textual descriptions, proving effective for fully synthetic images but struggling with partial manipulations that maintain semantic coherence with original captions.

ObjectFormer \cite{wang2022objectformer} represents a more advanced approach, utilizing a transformer-based architecture to fuse RGB and frequency features at the object level. While this enables more precise localization of manipulated regions compared to global classification methods, it remains vulnerable to instruction-guided edits that modify security-critical details while preserving object-level features. Furthermore, its architectural design lacks the capability to preserve spatial relationships between authentic and manipulated content, creating exploitable detection blind spots.

These multi-modal approaches, while advancing the state-of-art, fundamentally fail to address a critical security requirement: the ability to integrate information across modalities while preserving spatial hierarchies essential for detecting subtle, localized manipulations.

\subsection{Adversarial Robustness Analysis}
Recent security evaluations have exposed critical vulnerabilities in current deepfake detection systems. Abdullah et al. \cite{abdullahAnalysisRecentAdvances2024} demonstrated that state-of-the-art detectors suffer performance degradation of up to 53.92\% when confronted with user-customized generative models, highlighting a concerning lack of robustness against adaptive adversaries. Their work reveals a fundamental security trade-off: frequency-domain detectors demonstrate better generalization to unseen generation techniques but catastrophically fail under adversarial perturbations, while vision-language models show greater adversarial robustness but limited generalization capabilities.


This evaluation further demonstrated that foundation models enable semantic manipulations, such as altering specific object attributes, that successfully evade detection without introducing visible artifacts. These findings underscore the insufficiency of current detection paradigms against instruction-guided editing, which represents a natural evolution of these adversarial capabilities into user-friendly, automated workflows.

\subsection{Security Limitations in Spatial Understanding}
A fundamental security limitation in existing approaches is their inability to effectively capture and reason about spatial relationships between image elements, a critical requirement for detecting instruction guided edits, that manipulate specific objects or regions while preserving global context. Convolutional architectures \cite{heDeepResidualLearning2015, renFasterRCNNRealTime2016} lose spatial precision through pooling operations, while transformer-based approaches \cite{wang2022objectformer} struggle to maintain hierarchical spatial relationships despite their strong feature integration capabilities.

This architectural limitation creates a significant security vulnerability: existing detectors cannot effectively distinguish between legitimate variations in authentic images and subtle, security-critical manipulations that preserve most original content. Our proposed capsule-based approach directly addresses this vulnerability by encoding spatial hierarchies through dynamic routing mechanisms, enabling more effective detection of localized manipulations that preserve global image characteristics.

\subsection{Human Perception and Security Implications}
The security threat posed by instruction-guided image editing is amplified by its ability to produce manipulations that remain imperceptible to human observers while altering security-critical content. Recent perceptual studies demonstrate that humans detect less than 58\% of instruction-guided manipulations in controlled settings \cite{geSEEDDataEditTechnicalReport2024}, highlighting the critical need for automated detection systems that exceed human capabilities.

This perceptual limitation creates significant security implications for applications ranging from authentication systems to legal evidence evaluation, where manipulations below the threshold of human detection can nevertheless carry significant security consequences. Our work addresses this gap by providing detection capabilities specifically designed for manipulations that exploit human perceptual limitations.

\section{Background}

The rapid evolution of generative AI has introduced a new class of deepfakes that pose significant security challenges: \textbf{instruction-guided image edits}. Unlike traditional text-to-image generation, where the entire image is synthesized from scratch, instruction-guided editing targets real images and modifies specific semantic attributes (such as object identity, background context, or visual style) while preserving global visual coherence. These manipulations are particularly concerning from a cybersecurity standpoint because they maintain the illusion of authenticity while enabling adversaries to alter identity, fabricate visual evidence, or inject misinformation into trusted media pipelines.

\begin{figure}
    \centering
    \includegraphics[width=\linewidth]{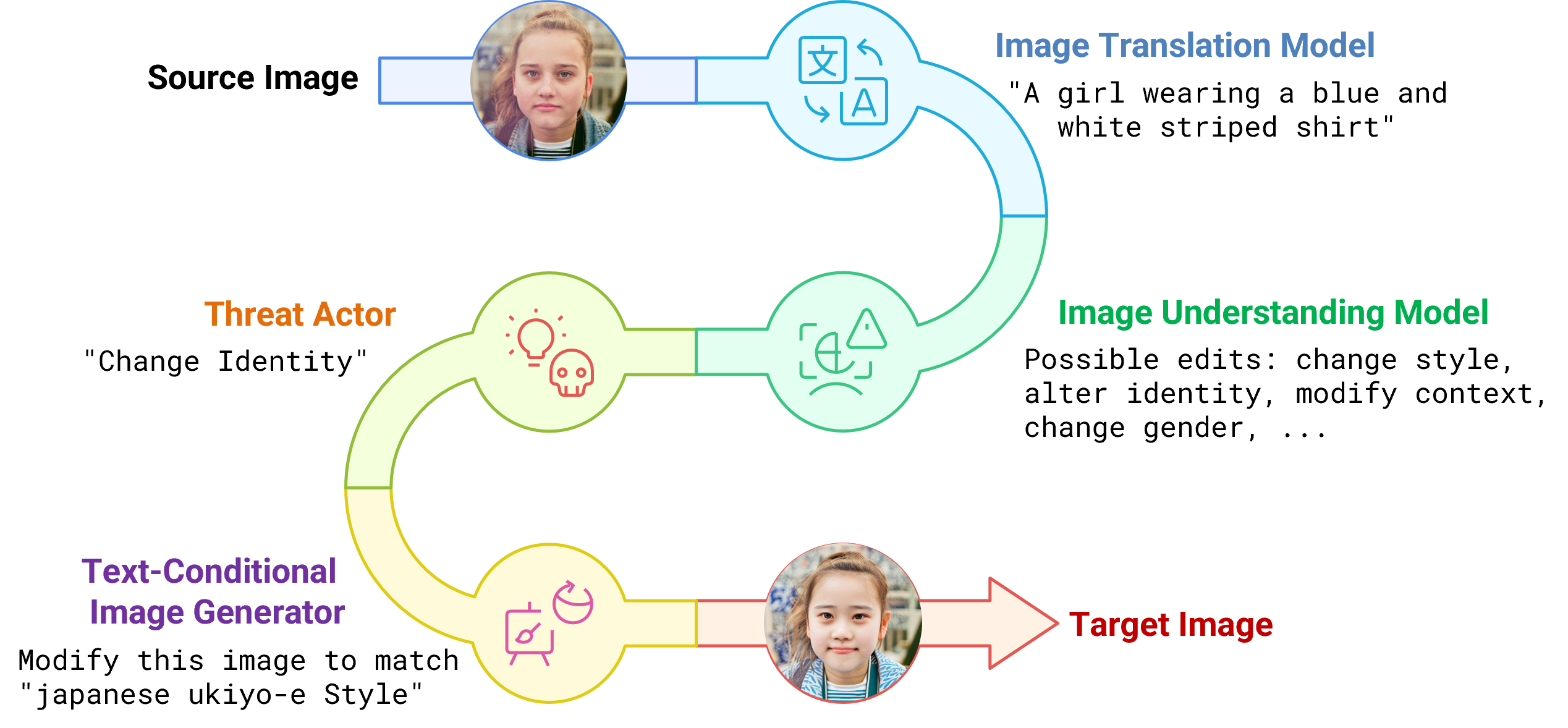}
    \caption{\textbf{Malicious Image Manipulation Pipeline.\label{fig:malicious_pipeline}} A threat actor uses generative AI tools to manipulate specific elements of an image, leveraging image translation and understanding models to guide semantic edits. These capabilities facilitate identity obfuscation, impersonation, and disinformation.}
\end{figure}

As illustrated in Figure~\ref{fig:malicious_pipeline}, the instruction-guided image editing pipeline comprises three key AI components, each playing a distinct role in enabling semantically precise and visually coherent manipulations.

First, an \textbf{image translation model} is used to convert the raw \textbf{source image} into a descriptive textual caption that semantically captures its visual content. This step, commonly implemented with models like CLIP \cite{radfordLearningTransferableVisual2021}, or BLIP-2 \cite{li2023BLIP2}, provides a language-based anchor that enables subsequent manipulation. For example, a facial image may be described as \textit{``a girl wearing a blue and white striped shirt''}, forming the basis for meaningful transformation prompts.

Next, an \textbf{image understanding model} analyzes the image and caption to infer actionable and context-aware manipulation instructions. These instructions may include commands such as \textit{``change identity''}, \textit{``alter expression''}, or \textit{``modify background context''}. Vision-language models like InstructBLIP \cite{dai2023instructblip} and LLaVA-1.5 \cite{liuImprovedBaselinesVisual} are commonly employed for this task, providing semantic reasoning over both visual and linguistic input to generate relevant, targeted editing prompts.

Finally, the generated instruction and original image are passed to a \textbf{text-conditional image generator}, such as InstructPix2Pix \cite{brooksInstructPix2PixLearningFollow2023} or PnP (Plug-and-Play Diffusion) \cite{tumanyanPlugandPlayDiffusionFeatures2022}. This model synthesizes the \textbf{target image} by applying the semantic transformation to the source image in a visually consistent manner. The generator is capable of modifying attributes like facial features, artistic style, or object presence while preserving background structure and global coherence, e.g., rendering the subject in \textit{``Japanese ukiyo-e style''} or removing/adding a specified object. The resulting \textbf{target image} appears coherent and visually consistent with the source, yet fundamentally altered in meaning or identity.

These instruction-driven modifications introduce an insidious \textit{attack vector}: they are easily accessible, require minimal technical expertise, and can evade both human perception and current deepfake detectors. Unlike end-to-end generated content, these manipulations \textbf{preserve the majority of the original image’s structure}, making the altered regions more difficult to isolate and detect. Moreover, such edits can be weaponized for \textit{identity obfuscation, digital impersonation, and misinformation dissemination}, particularly in settings involving biometric verification, surveillance imagery, or online social media platforms.

The security implications are profound. Traditional detectors often rely on global inconsistencies or synthetic generation artifacts, which are largely absent in these edits. Consequently, \textbf{instruction-guided image editing undermines the integrity of visual authentication systems}, calls into question the veracity of image-based evidence, and enables sophisticated adversaries to craft near-indistinguishable fake content at scale. In response to this emerging threat, it is essential to develop detection systems that go beyond pixel-level forensics and instead reason over \textbf{semantic, spatial, and modal inconsistencies}, a core motivation behind our proposed multimodal capsule-based approach.

\section{Our Approach: CapsFake}
\begin{figure*}
    \centering
    \resizebox{0.8\textwidth}{!}{%
        \includegraphics{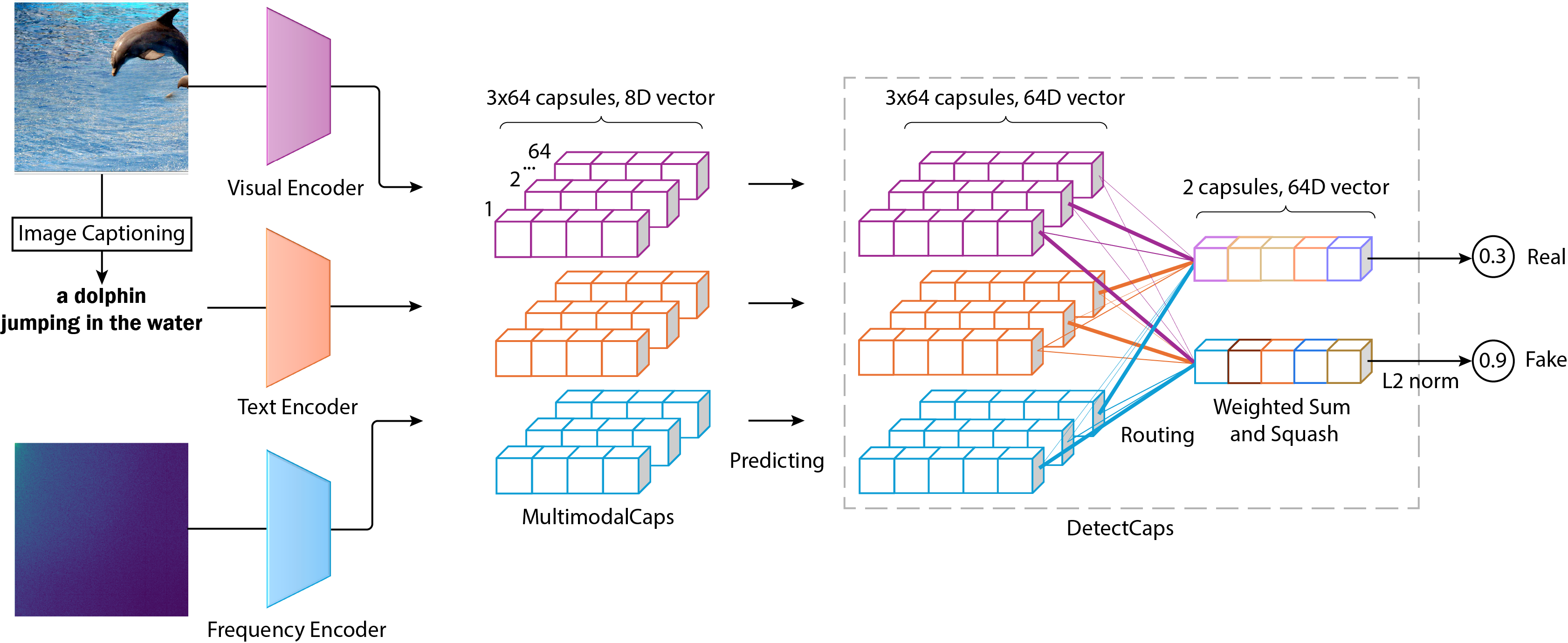}
    }
    \caption{Overview of the proposed multimodal capsule network, CapsFake, for detecting instruction-guided deepfakes.}
    \label{fig:framework}
\end{figure*}

\subsection{Overview}
Previous studies \cite{ojhaUniversalFakeImage2024, abdullahAnalysisRecentAdvances2024} highlight that single modalities exhibit distinct strengths in detecting deepfakes. Most existing methods rely on visual embeddings, extracting features from pretrained models such as CLIP-ViT \cite{radfordLearningTransferableVisual2021} or ConvNeXt \cite{liuConvNet2020s2022}. However, relying solely on the visual modality for detecting deepfakes in image editing is often insufficient. Detecting subtle deepfake artifacts within real images poses significant challenges, as these small changes are frequently obscured by the remaining regions that closely resemble the authentic image. This raises an important question: \textit{How can we develop an approach that effectively integrates information from multiple modalities to detect subtle manipulations in instructional image editing?} While each modality offers unique strengths, combining them into a unified framework is inherently challenging due to the differences in their characteristics. In this work, we successfully integrate features from visual, textual, and frequency modalities using a routing mechanism within a capsule network. Figure \ref{fig:framework} provides an overview of our proposed multimodal capsule network. The framework constructs multimodal capsules by combining low-level capsules from different modalities, which are subsequently used to predict high-level capsules representing real and manipulated entities. Details of the routing process are elaborated in Section \ref{sec:multimodal_capsule_network}. A key advantage of capsule networks is their ability to detect fine-grained patterns within an image. As the dimensionality of the high-level capsules increases, the network can capture a broader range of entity variations, enhancing its ability to effectively detect subtle changes and manipulations.

\subsection{Modality Embedding Extraction}
In this section, we elaborate on the feature extraction process for visual, textual, and frequency-domain embeddings utilized in our proposed framework.

\textbf{Visual Embeddings.} Following established baselines \cite{ojhaUniversalFakeImage2024,abdullahAnalysisRecentAdvances2024}, we utilize pretrained visual models to extract image features. Specifically, we adopt OpenCLIP-ConvNextLarge \cite{LaionCLIPconvnext_large_d_320laion2Bs29Bb131KftsoupHugging2023}, which has demonstrated strong performance across various downstream tasks. Unlike the original capsule network implementation \cite{sabourDynamicRoutingCapsules2017}, which employs shallow convolutional layers, we use pretrained embeddings as input to the capsule network. In deepfake image editing detection, shallow layers often fail to effectively capture subtle differences between images with minor changes. By contrast, features extracted by pretrained models provide richer representations, enabling better detection of fine-grained differences between real and manipulated images. We conduct experiments and discuss various design alternatives in Section \ref{sec:design_capsfake} in the Appendix.

\textbf{Text Embeddings.} Textual information plays a critical role as it provides semantic meaning for images. Previous study \cite{shaFAKEDetectionAttribution2023} shows that real images typically contain richer semantic content than fake images, which are often generated from simpler prompts. In image editing scenarios, the prompts associated with real and manipulated images can differ, as they describe specific edited attributes such as style, objects, or texture. To extract textual embeddings, we employ BLIP \cite{liBLIPBootstrappingLanguageImage2022} to generate captions for the images. This approach is particularly practical in real-world settings, where users often provide images without associated descriptive information.

\textbf{Frequency-Domain Embeddings.} Frequency-domain analysis has proven effective in identifying fake images, as generated images frequently exhibit artifacts or inconsistencies compared to real ones. In this work, we encapsulate frequency information within capsules, enabling different capsules to capture specific frequency ranges: low, medium, or high frequencies. This design allows high-level capsules to aggregate and select relevant frequency components, particularly those associated with manipulated areas. Inspired by \cite{rickerDetectionDiffusionModel2024}, we apply the Discrete Cosine Transform (DCT) to extract frequency-domain features
. This approach provides an additional channel of analysis to detect deepfake manipulations that may not be apparent in the spatial domain.

\subsection{Multimodal Capsule Routing Mechanism}
\label{sec:multimodal_capsule_network}
To detect subtle and adversarial image manipulations arising from instruction-guided deepfake editing, we introduce a novel multimodal capsule-based detection architecture. Our method is motivated by the growing sophistication of cross-modal generation pipelines \cite{zou2024cross, jimaging9060122} and the need for robust, context-aware forgery detection. The proposed framework integrates fine-grained cues from image content, textual instructions, and frequency-based analysis, using a shared capsule representation space coupled with a structured routing strategy.

Let \( \mathbf{X}_V, \mathbf{X}_T, \mathbf{X}_F \in \mathbb{R}^{N \times d} \) denote feature matrices extracted from the visual, textual, and frequency modalities, respectively. These embeddings represent the token-level descriptors of each modality and are obtained via modality-specific encoders such as vision transformers \cite{dosovitskiyImageWorth16x162021}, visual language models \cite{liBLIPBootstrappingLanguageImage2022}, or spectral CNNs \cite{rippel2015spectralCNNs}. To harmonize these representations in a unified capsule space, we apply separate linear projections:
\begin{align}
    \mathbf{C}_V &= \mathbf{X}_V \mathbf{W}_V, \\
    \mathbf{C}_T &= \mathbf{X}_T \mathbf{W}_T, \\
    \mathbf{C}_F &= \mathbf{X}_F \mathbf{W}_F,
\end{align}
where \( \mathbf{W}_V, \mathbf{W}_T, \mathbf{W}_F \in \mathbb{R}^{d \times d_i} \) are learnable projection matrices and \(d_i\) is the capsule dimension. This step preserves modality-specific characteristics while aligning their representations to a common space.

We concatenate the three projected matrices along the instance dimension to form a unified multimodal capsule input:
\begin{equation}
   \mathbf{C}_{\text{multi}} = \begin{bmatrix} \mathbf{C}_I \\ \mathbf{C}_T \\ \mathbf{C}_F \end{bmatrix} \in \mathbb{R}^{3N \times d_i}. 
\end{equation}

Each row of \( \mathbf{C}_{\text{multi}} \) serves as a low-level capsule embedding that will vote for higher-level semantic capsules representing real or manipulated content.

To capture hierarchical context, we map each low-level capsule to \(K\) high-level capsules through a shared learnable transformation tensor:
\begin{equation}
    \hat{\mathbf{U}} = \mathbf{C}_{\text{multi}} \mathbf{W}_{\text{route}},
\end{equation}

where \( \mathbf{W}_{\text{route}} \in \mathbb{R}^{d_i \times K \times d_k} \) defines a set of transformation matrices. The output \( \hat{\mathbf{U}} \in \mathbb{R}^{3N \times K \times d_k} \) contains predictions \( \hat{\mathbf{U}}_{i,k} \) from each input capsule \(i\) to output capsule \(k\).

The agreement between capsules is mediated by dynamic routing weights \( \alpha_{i,k} \), computed via softmax-normalized logits \( a_{i,k} \):
\begin{equation}
    \alpha_{i,k} = \frac{\exp(a_{i,k})}{\sum_{j=1}^{K} \exp(a_{i,j})}.
\end{equation}

These weights determine the influence of each low-level capsule on high-level capsules. Using these weights, we compute the aggregated input to each high-level capsule as:
\begin{equation}
\mathbf{s}_k = \sum_{i=1}^{3N} \alpha_{i,k} \hat{\mathbf{U}}_{i,k}.
\end{equation}
To ensure vector norms are bounded and semantically meaningful, we apply a squashing function:
\begin{equation}
\mathbf{v}_k = \frac{\|\mathbf{s}_k\|^2}{1 + \|\mathbf{s}_k\|^2} \cdot \frac{\mathbf{s}_k}{\|\mathbf{s}_k\|},
\end{equation}
where \( \mathbf{v}_k \) serves as the final embedding for class capsule \(k\).

The routing logits \( a_{i,k} \) are iteratively refined using the degree of alignment between predicted and actual capsule outputs:
\begin{equation}
a_{i,k} \leftarrow a_{i,k} + \langle \hat{\mathbf{U}}_{i,k}, \mathbf{v}_k \rangle.
\end{equation}
This update mechanism encourages coherent routing paths by amplifying strong agreements and attenuating weak ones.

\subsection{Training Objective}
We adopt a max-margin loss that encourages high confidence for the correct class while penalizing uncertain or incorrect predictions:
\begin{equation}
\mathcal{L} = \sum_k y_k \cdot \max(0, \nu - \|\mathbf{v}_k\|)^2 + \lambda \cdot (1 - y_k) \cdot \max(0, \|\mathbf{v}_k\| - \mu)^2,
\end{equation}
where \( y_k \in \{0,1\} \), \( \mu, \nu \) are margin thresholds, and \( \lambda \) down-weights the loss of the negative class. The use of dynamic routing combined with a max-margin loss serves several security-oriented purposes:  
(i) the routing mechanism can filter out inconsistent or noisy feature contributions, making it more difficult for adversarial perturbations to influence capsule outputs;  
(ii) the model dynamically adapts to diverse manipulation styles by learning the transformation matrix $\mathbf{W}_{\text{route}}$ across modalities, which enhances CapsFake's robustness to different editing styles and manipulation patterns; and  
(iii) the magnitude and direction of capsule vectors provide interpretable representations of semantic consistency and the presence of manipulated objects, enabling saliency-based visualization and fine-grained localization of edits. 

Finally, the complete routing procedure across multimodal capsules is detailed in Algorithm~\ref{alg:multimodal_capsules}.








\begin{algorithm}

\caption{Multimodal Capsule Routing\label{alg:multimodal_capsules}}
\begin{algorithmic}[1]

\REQUIRE Capsule input matrix $\mathbf{C}_{\text{multi}} \in \mathbb{R}^{3N \times d_i}$, routing iterations $R$, number of output capsules $K$
\ENSURE Final output capsule vectors $\{\mathbf{v}_k\}_{k=1}^K$, where each $\mathbf{v}_k \in \mathbb{R}^{d_k}$ represents the routed embedding of class capsule $k$
\STATE Initialize logits $a_{i,k} \leftarrow 0$ for all $i,k$
\FOR{$r = 1$ to $R$}
    \STATE Compute weights: $\alpha_{i,k} = \text{softmax}_k(a_{i,k})$
    \STATE Predict outputs: $\hat{\mathbf{U}}_{i,k} = \mathbf{C}_{\text{multi}}^{(i)} \cdot \mathbf{W}_{\text{route}}^{(:,k,:)}$
    \STATE Aggregate: $\mathbf{s}_k = \sum_{i} \alpha_{i,k} \hat{\mathbf{U}}_{i,k}$
    \STATE Squash: $\mathbf{v}_k = \frac{\|\mathbf{s}_k\|^2}{1 + \|\mathbf{s}_k\|^2} \cdot \frac{\mathbf{s}_k}{\|\mathbf{s}_k\|}$
    \STATE Update logits: $a_{i,k} \leftarrow a_{i,k} + \langle \hat{\mathbf{U}}_{i,k}, \mathbf{v}_k \rangle$
\ENDFOR

\end{algorithmic}
\end{algorithm}

\section{Deepfake Image Editing Detection}

\subsection{Dataset construction}\label{sec:dataset_construction}
We evaluate our proposed method on a diverse set of datasets, including MagicBrush, Unsplash Edits, Open Images Edits, and Multi-turn Edits. These datasets span natural scenes, everyday objects, and personal identities involving human faces, providing a comprehensive benchmark for semantic deepfake detection. While facial manipulation detection has been extensively studied~\cite{rosslerFaceForensicsLearningDetect2019,guoFakeFaceDetection2020,dangDetectionDigitalFace2020,reissDetectingDeepfakesSeeing2023,Rehaan02012024}, our evaluation broadens the scope to include complex, multi-entity visual compositions. This diversity introduces new challenges for detection systems, particularly in identifying subtle, instruction-guided edits in context-rich environments.

\subsubsection{MagicBrush}
The dataset consists of edits applied to 80 different objects from the MS COCO dataset \cite{linMicrosoftCOCOCommon2015}. Images were manually annotated by creating text prompts and drawing masks, after which the DALL-E 2 platform \cite{DALLE22022} was utilized for inpainting. The MagicBrush dataset comprises a total of 10,626 images, evenly divided into 5,313 source (clean) images and 5,313 target (edited) images. MagicBrush features a diverse set of edit instructions, including adding, removing, or replacing objects; modifying actions; altering colors; editing text or patterns; among others. To evaluate the performance of our model for deepfake detection, the dataset was split into training, validation, and test sets, containing 9,024; 532; and 1,070 images, respectively.

\subsubsection{Unsplash Edits and Open Images Edits}
These datasets are derived from the first part of SEED-Data-Edit \cite{geSEEDDataEditTechnicalReport2024}, consisting of 30,000 images from Unsplash and 52,144 images from Open Images. The original images were sourced from their respective datasets. We utilize image edits based on instruction-guided manipulation for both the Unsplash \cite{zahidalichesserlukeandcarbonetimothyUnsplash2023} and Open Images \cite{kuznetsovaOpenImagesDataset2020} datasets. For the Unsplash Edits dataset, the images are divided into 24,000 for training, 3,000 for validation, and 3,000 for testing. Similarly, the Open Images Edits dataset is partitioned into 41,714 training, 5,214 validation, and 5,216 testing images.

\subsubsection{Multi-turn Edits}
This dataset addresses the case where manipulated regions are generated through skilled photo editing involving multiple editing rounds. Images were collected from Unsplash \cite{zahidalichesserlukeandcarbonetimothyUnsplash2023}, SAM \cite{kirillovSegmentAnything2023}, and JourneyDB \cite{sunJourneyDBBenchmarkGenerative2023}. We randomly selected a subset of 140,966 images with a variety of edits applied to different regions. The edits were performed over five rounds, where Photoshop experts applied modifications such as removing, adding, or replacing objects; changing actions; altering text or patterns; or modifying the count of objects. Unlike MagicBrush, Unsplash Edits, and Open Images Edits, which involve single-turn edits, this dataset allows us to evaluate our model’s performance on multi-round edits. Additionally, we analyze how detection performance varies across 1 to 5 rounds of edits in Section \ref{sec:multi-turn-edits}. Due to space constraints, we present results for Three-turn Edits here. The results for One-turn Edits, Two-turn Edits, Four-turn Edits, and Five-turn Edits are provided in Tables \ref{tab:1-turn}, \ref{tab:2-turn}, \ref{tab:4-turn}, and \ref{tab:5-turn} in the Appendix.

Detailed information on the data split and image sizes is provided in Table \ref{tab:dataset-details} (Appendix \ref{sec:dataset_information}). 


\begin{table}
\centering
\caption{Datasets used for main experiments.}\label{tab:dataset_info}
\resizebox{\columnwidth}{!}{%
\begin{tabular}{p{2.8cm}ccc}
\hline
Dataset & Type & Total Images & Image Size \\ \hline
\multirow{2}{*}{MagicBrush}
   & Source & 5,313 & 500 $\times$ 500 \\
   & Target & 5,313 & 1024 $\times$ 1024 \\ \hline
\multirow{2}{*}{Unsplash Edits}
   & Source & 15,000 & 512 $\times$ 512 \\
   & Target & 15,000 & 512 $\times$ 512 \\ \hline
\multirow{2}{*}{Open Images Edits}
   & Source & 26,072 & 768 $\times$ 768 \\
   & Target & 26,072 & 768 $\times$ 768 \\ \hline
\multirow{2}{*}{One-turn Edits}
   & Source & 18,166 & Varies \\
   & Target & 18,166 & Varies \\ \hline
\multirow{2}{*}{Two-turn Edits}
   & Source & 18,555 & Varies \\
   & Target & 18,555 & Varies \\ \hline
\multirow{2}{*}{Three-turn Edits}
   & Source & 18,173 & Varies \\
   & Target & 18,173 & Varies \\ \hline
\multirow{2}{*}{Four-turn Edits}
   & Source & 11,560 & Varies \\
   & Target & 11,560 & Varies \\ \hline
\multirow{2}{*}{Five-turn Edits}
   & Source & 4,029 & Varies \\
   & Target & 4,029 & Varies \\ \hline
\end{tabular}
}
\end{table}

\subsection{Implementation Details}\label{sec:implementation_details}

Input images are resized to $320 \times 320$ pixels to meet the requirements of the pretrained OpenCLIP-ConvNextLarge model \cite{radfordLearningTransferableVisual2021}, which serves as the encoder for both visual and textual modalities. The corresponding text instructions are processed by the same OpenCLIP model to extract sentence-level embeddings. For the frequency modality, we compute the Discrete Cosine Transform (DCT) on the input image, taking the logarithm of the absolute DCT coefficients. This is followed by normalization (mean subtraction and standard deviation division) and projection into the shared capsule space.

The extracted embeddings from visual, text, and frequency domains are each 768-dimensional ($d = 768$). These embeddings are then passed through a capsule encoder module. Each module outputs an 8-dimensional vector, resulting in $N = 64$ capsules per modality with capsule dimension $d_i = 8$. Therefore, the unified multimodal capsule input matrix $\mathbf{C}_{\text{multi}} \in \mathbb{R}^{192 \times 8}$ contains $3N = 192$ low-level capsules across all modalities.

To enable capsule-specific transformation, we learn a unique matrix $\mathbf{W}_{i,k} \in \mathbb{R}^{8 \times 64}$ for each pair of input capsule $i$ and output capsule $k$. The total number of output capsules is $K = 2$, each with dimensionality $d_k = 64$. As a result, the routing transformation tensor $\mathbf{W}_{\text{route}}$ encodes all such transformations. Dynamic routing is performed over $R = 3$ iterations to refine coupling coefficients $\alpha_{i,k}$ based on the agreement between predictions and outputs.

For training, we use the AdamW optimizer \cite{loshchilov2018decoupled} with a learning rate of $1 \times 10^{-4}$ and a batch size of 64. The model is trained for 30 epochs with early stopping based on validation accuracy. All experiments are conducted using 8 NVIDIA Tesla V100-SXM2 GPUs, each with 32GB of VRAM. For baseline comparisons, we adopted publicly available official implementations and preserved all hyperparameter settings as per their original configurations.

\begin{table*}[t]
\begin{minipage}{0.48\textwidth}
\centering
\caption{Deepfake detection performance on MagicBrush.}\label{tab:main_magic}
\begin{tabular}{ccccc}
\hline
Methods & Precision & Recall & F1 & Accuracy\tabularnewline
\hline
UnivCLIP & 69.78 & 71.21 & 70.49 & 70.19\tabularnewline
DCT & 98.85 & 96.26 & 97.54 & 97.57\tabularnewline
ZeroFake & 53.02 & 61.68 & 57.29 & 54.02\tabularnewline
DE-FAKE & 79.81 & 92.34 & 85.62 & 84.49\tabularnewline
ObjectFormer & 73.55 & 92.52 & 81.95 & 79.65\tabularnewline
UnivConv2B & \textbf{100.0} & 94.58 & 97.21 & 97.29\tabularnewline
\hline
\textbf{CapsFake} & 98.71 & \textbf{100.0} & \textbf{99.35} & \textbf{99.35}\tabularnewline
\hline
\end{tabular}
\end{minipage}
\hfill
\begin{minipage}{0.48\textwidth}
\centering
\caption{Deepfake detection performance on Unsplash Edits.}\label{tab:main_unsplash}
\begin{tabular}{ccccc}
\hline
Methods & Precision & Recall & F1 & Accuracy\tabularnewline
\hline
UnivCLIP & 85.18 & 85.80 & 85.49 & 85.43\tabularnewline
DCT & 88.06 & 71.80 & 79.10 & 81.03\tabularnewline
ZeroFake & 56.66 & 55.33 & 55.99 & 56.50\tabularnewline
DE-FAKE & 89.53 & 76.93 & 82.75 & 83.97\tabularnewline
ObjectFormer & 81.74 & 97.27 & 88.83 & 87.77\tabularnewline
UnivConv2B & 93.98 & 95.80 & 94.88 & 94.83\tabularnewline
\hline
\textbf{CapsFake} & \textbf{99.20} & \textbf{99.80} & \textbf{99.50} & \textbf{99.50}\tabularnewline
\hline
\end{tabular}
\end{minipage}

\vspace{1em}

\begin{minipage}{0.48\textwidth}
\centering
\caption{Deepfake detection performance on Open Images Edits.}\label{tab:main_open}
\begin{tabular}{ccccc}
\hline
Methods & Precision & Recall & F1 & Accuracy\tabularnewline
\hline
UnivCLIP & 71.75 & 53.93 & 61.58 & 66.35\tabularnewline
DCT & 77.04 & 41.68 & 54.09 & 64.63\tabularnewline
ZeroFake & 57.53 & 27.11 & 36.85 & 53.55\tabularnewline
DE-FAKE & 72.03 & 46.32 & 56.38 & 64.17\tabularnewline
ObjectFormer & 79.63 & 58.32 & 67.33 & 71.71\tabularnewline
UnivConv2B & 63.58 & 77.65 & 69.91 & 66.58\tabularnewline
\hline
\textbf{CapsFake} & \textbf{89.98} & \textbf{95.74} & \textbf{92.77} & \textbf{92.54}\tabularnewline
\hline
\end{tabular}
\end{minipage}
\hfill
\begin{minipage}{0.48\textwidth}
\centering
\caption{Deepfake detection performance on Three-turn Edits.}\label{tab:main_photoshop}
\begin{tabular}{ccccc}
\hline
Methods & Precision & Recall & F1 & Accuracy\tabularnewline
\hline
UnivCLIP & 77.11 & 65.40 & 70.77 & 72.99\tabularnewline
DCT & 57.23 & 62.93 & 59.94 & 57.95\tabularnewline
ZeroFake & 56.67 & 38.56 & 45.89 & 54.54\tabularnewline
DE-FAKE & 81.01 & 62.16 & 70.34 & 73.79\tabularnewline
ObjectFormer & 75.37 & 63.81 & 69.11 & 71.48\tabularnewline
UnivConv2B & 75.10 & 85.75 & 80.07 & 78.66\tabularnewline
\hline
\textbf{CapsFake} & \textbf{95.38} & \textbf{90.92} & \textbf{90.92} & \textbf{93.26}\tabularnewline
\hline
\end{tabular}
\end{minipage}
\end{table*}

\subsection{Results and Discussion}
\label{Results and Discussion}
We compare the performance of our proposed method, CapsFake, 
with state-of-the-art deepfake detection models: DE-FAKE \cite{shaFAKEDetectionAttribution2023}, ZeroFake \cite{shaZeroFakeZeroShotDetection}, DCT \cite{rickerDetectionDiffusionModel2024}, UnivCLIP \cite{ojhaUniversalFakeImage2024}, ObjectFormer \cite{wang2022objectformer} and UnivConv2B \cite{abdullahAnalysisRecentAdvances2024}, as discussed in Section \ref{sec:related_work}. The evaluation metrics include precision, recall, F1 score, and accuracy. We evaluate these methods on datasets of increasing difficulty, ranging from single-turn edits, including MagicBrush, Unsplash Edits, and Open Images Edits, to Multi-turn Edits involving multiple rounds of editing. All baseline models are trained from scratch using their official GitHub implementations. We start from the authors’ recommended settings and further tune hyperparameters to ensure optimal performance and fair comparison.

Table~\ref{tab:main_magic} reports the deepfake detection performance on the MagicBrush dataset. UnivConv2B achieved the highest precision (100\%), indicating strong reliability in identifying authentic content. CapsFake achieved perfect recall (100\%) and the highest F1 score (99.35\%), suggesting effective coverage in identifying manipulated instances. ObjectFormer also performed competitively, with an F1 score of 81.95\% driven by strong recall (92.52\%) and moderate precision (73.55\%). Compared to DE-FAKE, ObjectFormer achieved a marginally higher recall but slightly lower F1 due to decreased precision. Overall, the MagicBrush dataset, comprising large-area inpainting, provides relatively clear signals for detection models, resulting in higher absolute scores across multiple methods.

Table~\ref{tab:main_unsplash} presents results on the Unsplash Edits dataset. CapsFake attained the highest performance across all metrics, including an F1 score of 99.50\%. ObjectFormer ranked second in both F1 (88.83\%) and recall (97.27\%), highlighting its robustness to diverse edit types. In comparison to DE-FAKE, ObjectFormer achieved a 6.08\% improvement in F1 score. UnivConv2B maintained strong results, particularly in precision and recall, while UnivCLIP and DCT showed moderate performance. These results reflect the relatively complex and stylistic nature of edits in this dataset.

Tables~\ref{tab:main_open} and~\ref{tab:main_photoshop} report performance on Open Images Edits and Three-turn Edits, both of which involve more subtle or localized manipulations. All models experienced a relative drop in overall scores on these challenging benchmarks. On Open Images Edits, ObjectFormer achieved 67.33\% F1, outperforming DE-FAKE and UnivCLIP by 10.95\% and 5.75\%, respectively. On Three-turn Edits, ObjectFormer achieved 69.11\% F1, ranking above DE-FAKE and ZeroFake. UnivConv2B and CapsFake yielded the highest overall scores across all metrics in both datasets, with CapsFake achieving an F1 score of 92.77\% and 90.92\% on Open Images Edits and Three-turn Edits, respectively. These findings highlight the ability of certain models to scale to more complex manipulation scenarios and demonstrate the potential benefits of multimodal fusion and capsule-based reasoning for robust detection.

\section{Robustness Evaluation}

\begin{figure*}
\begin{centering}
\resizebox{1.0\textwidth}{!}{%
\includegraphics{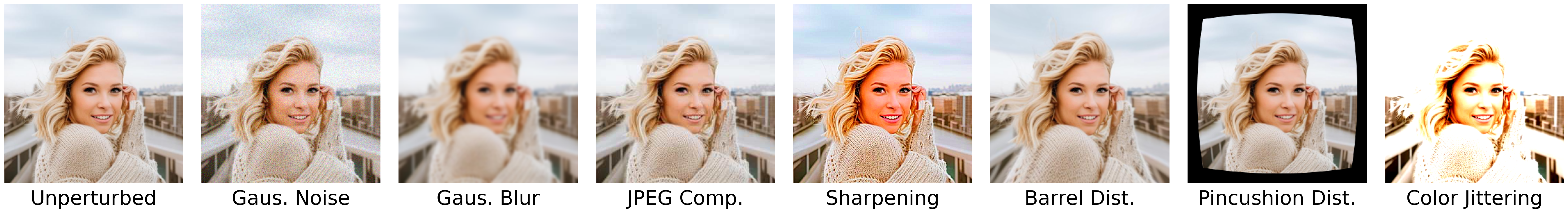}
}
\vspace{-5mm}
\caption{Examples of various natural perturbations analyzed in our study.}
\label{fig:natural_perturbation}
\par\end{centering}
\end{figure*}

\begin{table*}
\centering{}\caption{Average F1 scores for varying intensities of natural perturbations on manipulated images from the Unsplash dataset.}
\label{tab:abla_natural_pertub}
\resizebox{1.0\textwidth}{!}{%
\begin{tabular}{ccccccccc}
\hline
Methods & Unperturbed & Gaus. Noise & Gaus. Blur & JPEG Comp. & Sharpening & Barrel Dist. & Pincushion Dist. & Color Jittering \tabularnewline
\hline
UnivCLIP & 85.49 & 81.05 & 81.85 & 83.47 & 83.76 & 82.57 & 83.16 & 78.06 \tabularnewline
DCT & 79.10 & 66.91 & 44.96 & 52.80 & 69.37 & 09.16 & 21.73 & 64.34 \tabularnewline
DE-FAKE & 82.75 & 69.64 & 58.24 & 76.20 & 84.30 & 79.05 & 57.27 & 68.55 \tabularnewline
ObjectFormer & 88.83 & 07.16 & 40.90 & 42.04 & 67.44 & 68.79 & 68.10 & 54.39 \tabularnewline
UnivConv2B & 94.88 & 86.58 & 79.75 & 88.73 & 78.63 & 88.22 & 94.88 & 80.36 \tabularnewline
\hline
\textbf{CapsFake} & \textbf{99.50} & \textbf{97.20} & \textbf{96.81} & \textbf{94.48} & \textbf{95.56} & \textbf{99.04} & \textbf{99.48} & \textbf{99.17} \tabularnewline
\hline
\end{tabular}}
\end{table*}

\subsection{Natural Perturbations}
We evaluate CapsFake under a suite of common image perturbations to assess its robustness in realistic deployment scenarios. Using the manipulated portion of the Unsplash dataset~\cite{zahidalichesserlukeandcarbonetimothyUnsplash2023}, we apply transformations such as Gaussian noise, blur, JPEG compression, sharpening, lens distortions, and color jittering, each at multiple intensities. These transformations simulate natural variations (e.g., user edits, compression artifacts) that are not semantically manipulative but may cause misclassification in sensitive detection systems.

\textbf{Gaussian Noise and Blur.} Gaussian noise is added with increasing standard deviations $\sigma \in \{0.01, 0.0125, 0.025, 0.05, 0.1\}$, while blur is applied with $\sigma \in \{0.5, 1.0, 2.0, 3.0, 4.0\}$.

\textbf{JPEG Compression.} Quality factors from $\{10, 30, 50, 70, 90\}$ are used to test the model’s tolerance to lossy compression~\cite{Lau2003,s23073400}.

\textbf{Sharpening.} Image contrast is enhanced across sharpening factors $\in \{0.1, 0.5, 1.0, 1.5, 2.0\}$.

\textbf{Optical Distortion.} Barrel and pincushion distortions are introduced using radial coefficients $\kappa \in \{-0.1, \dots, -0.5\}$ and $\{0.1, \dots, 0.5\}$ respectively, following~\cite{johnstonREALTIMEFPGAIMPLEMENTATION}.

\textbf{Color Jittering.} Brightness and contrast are adjusted using scale factors $\in \{0.75, 1.25, 1.5, 2.0\}$.

For each transformation, we report average F1 scores over intensity levels. These experiments are critical for evaluating resilience to non-adversarial perturbations that may still impact system reliability. An illustration of natural perturbation effects is shown in Figure~\ref{fig:natural_perturbation}.


Table~\ref{tab:abla_natural_pertub} shows that CapsFake achieves the highest average F1 scores across all natural perturbations, maintaining robust performance above 94\% in every scenario (e.g., 97.20\% under Gaussian noise, 96.81\% under blur, 99.48\% under pincushion distortion). In contrast, UnivConv2B performs well overall (94.88\% unperturbed) but drops to 79.75\% under blur and 80.36\% under color jittering. UnivCLIP is stable on moderate distortions (e.g., 83.47\% JPEG) but weaker under lens effects (82.57\% barrel, 83.16\% pincushion). ObjectFormer is highly vulnerable to noise (7.16\%) and blur (40.90\%), indicating limited resilience. DCT and DE-FAKE show significant sensitivity to low-level perturbations, especially blur (44.96\% and 58.24\%, respectively). These results demonstrate that CapsFake’s multimodal capsule routing offers strong robustness under realistic degradation, outperforming all baselines under every tested condition.

\subsection{White-Box Adversarial Attacks}

We evaluate the robustness of our model against two widely used adversarial attacks: the Fast Gradient Sign Method (FGSM)~\cite{goodfellowExplainingHarnessingAdversarial2015} and Projected Gradient Descent (PGD)~\cite{madryDeepLearningModels2019}. These attacks are conducted in a white-box setting, where the attacker has complete knowledge of the detection models, including its architecture, parameters, and training loss functions of detectors. This setup simulates the worst-case scenario for robustness evaluation. Let $\bx$ denote an input image and $y$ the corresponding ground truth label (e.g., fake or real). The function $f(\bx; \phi)$ represents the prediction of a detector model $f$ parameterized by $\phi$, and $\mathcal{L}(f(\bx;\phi), y)$ is the loss function corresponding to its detector (e.g., cross-entropy loss). The adversarial image is denoted by $\bx_{\text{adv}}$.

\noindent\textbf{Attack Motivation.} Adversarial attacks aim to generate small, human-imperceptible perturbations that significantly degrade the model's performance. By evaluating models under such attacks, we assess their ability to maintain high recall when faced with manipulated inputs that attempt to bypass detection.

\vspace{0.5em}
\noindent\textbf{FGSM Attack.} FGSM perturbs the input image in the direction of the gradient of the loss function with respect to the input, causing maximal increase in loss with minimal perturbation. The adversarial example is computed as:
\begin{equation}
    \bx_{\text{adv}} = \bx + \eta \cdot \text{sign}(\nabla_\bx \mathcal{L}(f(\bx;\phi), y)),
\end{equation}
where $\eta$ is the perturbation magnitude. In our experiments, we use $\eta = 0.005$.

\vspace{0.5em}
\noindent\textbf{PGD Attack.\label{par:pgd}} PGD extends FGSM by applying iterative perturbations, each followed by projection back into the $\epsilon$-ball centered at the original image. This makes PGD a stronger and more effective white-box attack. The iterative update is given by:
\begin{equation}
    \bx^{t+1}_{\text{adv}} = \text{Proj}_{B_{\epsilon}(\bx)} \left( \bx^t_{\text{adv}} + \gamma \cdot \text{sign}(\nabla_\bx \mathcal{L}(f(\bx^t_{\text{adv}};\phi), y)) \right),
\end{equation}
where $\bx^0_{\text{adv}} = \bx$, $\epsilon = 0.005$ is the maximum perturbation bound, $\gamma = 0.008$ is the step size, and $t = 10$ is the number of iterations.

\vspace{0.5em}
\noindent\textbf{Evaluation Setup.} Both attacks are applied to fake images in the MagicBrush test set, which demonstrates strong recall on clean (natural) images. We report recall scores on clean inputs (natural accuracy) and adversarially perturbed inputs (attack accuracy). The goal is to assess the resilience of each detector model to adversarial manipulation targeting fake image detection.

\begin{table}
\centering{}\caption{Detection performance (Recall score) against adversarial attacks on MagicBrush.}\label{tab:abla_adv_attack}
\begin{tabular}{cccc}
\hline
Methods & Natural accuracy & FGSM & PGD\tabularnewline
\hline
UnivCLIP & 71.21 & 63.81 & 70.84\tabularnewline
DCT & 96.26 & 88.41 & 90.28\tabularnewline
DE-FAKE & 92.34 & 13.83 & 17.38\tabularnewline
ObjectFormer & 92.52 & 91.59 & 80.93\tabularnewline
UnivConv2B & 94.58 & 15.33 & 51.03\tabularnewline
\hline
\textbf{CapsFake} & \textbf{100.0} & \textbf{100.0} & \textbf{96.82}\tabularnewline
\hline
\end{tabular}
\end{table}

Table \ref{tab:abla_adv_attack} presents the detection performance (recall scores) against adversarial attacks. While state-of-the-art detectors struggle to maintain high recall scores under adversarial perturbations, CapsFake achieves exceptional robustness, with 100\% recall on both natural and adversarial images for FGSM attacks and 96.82\% for PGD attacks, significantly outperforming other methods. For instance, UnivConv2B experiences a substantial drop in performance under adversarial attacks, with recall scores decreasing from 94.28\% on clean images to 15.33\% under FGSM and 51.03\% under PGD. This vulnerability arises because visual modalities are particularly sensitive to adversarial perturbations \cite{abdullahAnalysisRecentAdvances2024}. Models focusing on the frequency domain, such as the DCT-based model, demonstrate better resistance, with only minor reductions in recall under FGSM and PGD attacks. In contrast, CapsFake exhibits superior robustness, which can be attributed to its multimodal capsule architecture. Multimodal capsules encode rich representations of the image, and the real and manipulated capsules in the DetectCaps layer preserve key distinguishing features between real and fake images. Through the routing mechanism, these capsules filter out inconsistent or noisy information, effectively mitigating the impact of adversarial perturbations and ensuring robust performance across a wide range of attacks.

\begin{table}
\centering
\caption{CapsFake recall (\%) under transferred adversarial examples.}\label{tab:adv_transfer_grouped}
\resizebox{\columnwidth}{!}{%
\begin{tabular}{llc}
\toprule 
Attack Type & Source Detector & CapsFake Recall (\%) \\
\midrule 
\multirow{5}{*}{PGD Attack}
    & DCT & 100.00 \\
    & DE-FAKE & 100.00 \\
    & UniCLIP & 99.44 \\
    & ObjectFormer & 99.25 \\
    & UnivConv2B & 91.78 \\
\midrule 
\multirow{3}{*}{Semantic Attack}
    & CLIP-ResNet & 99.70 \\
    & EfficientNet & 99.90 \\
    & ViT & 100.00 \\
\bottomrule 
\end{tabular}
}%
\end{table}

\subsection{Black-Box Transfer Attacks}
In this setting, we assume the attacker has no knowledge of the CapsFake architecture, including its capsule-based design or training parameters. Instead, the adversary crafts inputs using surrogate detectors to generate adversarial examples, which are then transferred to CapsFake in an attempt to bypass detection.

We evaluate two categories of black-box transfer attacks: (i) \textit{PGD-based adversarial attacks}, as described in Section~\ref{par:pgd}, where perturbations are crafted using baseline detectors such as DCT, DE-FAKE, and UniCLIP on the MagicBrush dataset; and (ii) \textit{semantic attacks} as introduced in~\cite{abdullahAnalysisRecentAdvances2024}, where facial images are manipulated using StyleCLIP dataset~\cite{patashnikStyleCLIPTextDrivenManipulation2021a}. In the semantic setting, attackers utilize large-scale vision backbones including CLIP-ResNet~\cite{radfordLearningTransferableVisual2021}, EfficientNet~\cite{tan2019efficientnet}, and ViT~\cite{dosovitskiyImageWorth16x162021}, each conditioned on text prompts such as \textit{``a smiling face,”} \textit{``a face with lipstick,”} or \textit{``a face with glasses” }to alter semantic attributes of facial images without introducing perceptible distortions.

Table~\ref{tab:adv_transfer_grouped} reports the recall of CapsFake under these black-box transfer scenarios. Against PGD attacks transferred from state-of-the-art detectors, CapsFake demonstrates strong robustness, achieving 100.00\% recall when adversaries are generated by DCT and DE-FAKE, and maintaining high scores against UniCLIP (99.44\%), ObjectFormer (99.25\%), and UnivConv2B (91.78\%). Notably, under semantic attacks, where manipulations are visually coherent yet semantically altered, CapsFake maintains equally strong performance, achieving 99.70\% recall with CLIP-ResNet, 99.90\% with EfficientNet, and 100.00\% with ViT. These results demonstrate that CapsFake generalizes well to unseen manipulations, both in the form of gradient-based attacks and semantically guided edits, further validating the model's robustness and cross-modal consistency in adversarial environments.

\subsection{Robustness to Progressive Multi-turn Edits}
\label{sec:multi-turn-edits}

\begin{figure}[ht]
    \centering
    \resizebox{0.7\columnwidth}{!}{%
        \includegraphics{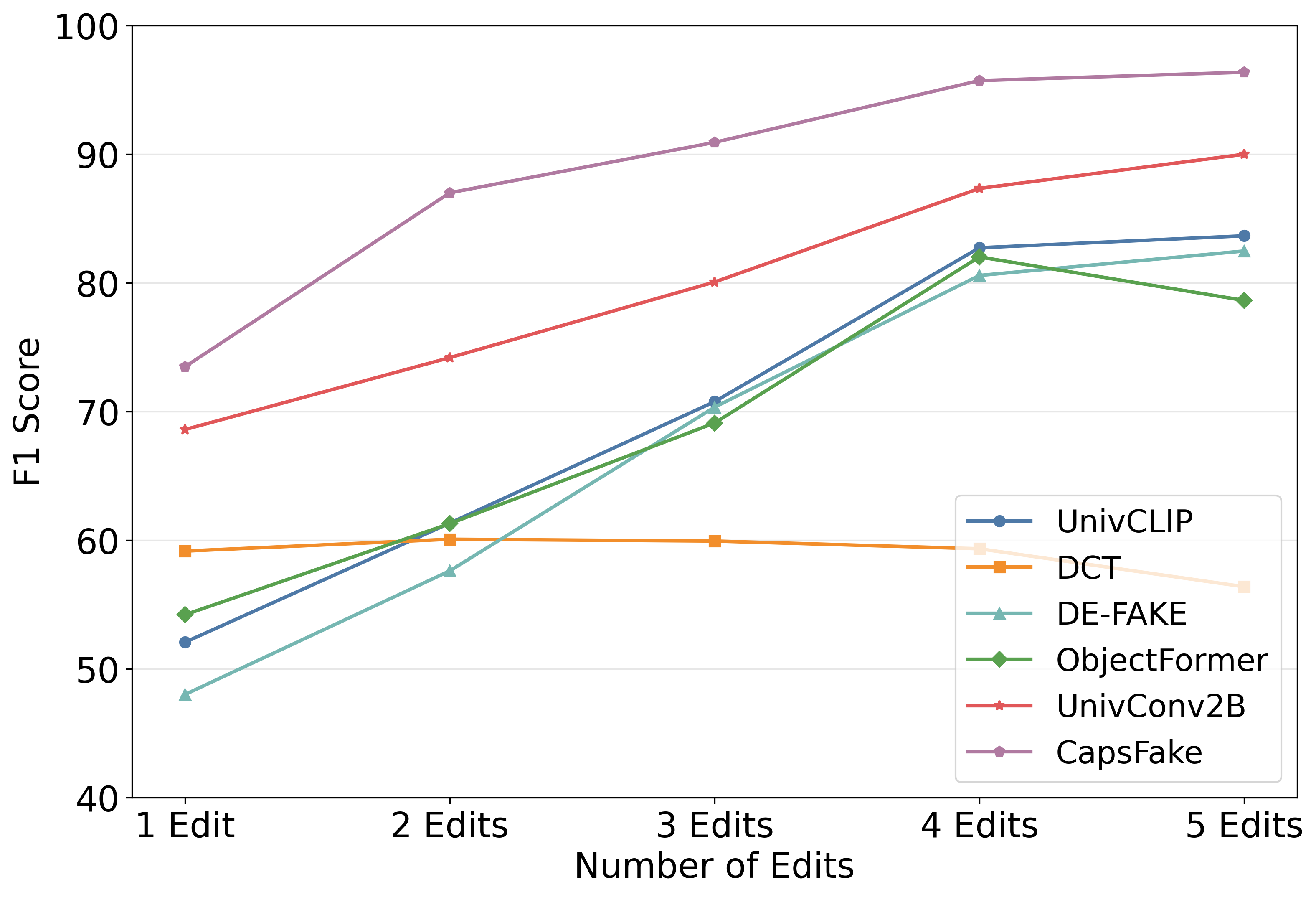}
    }
    \caption{Detection performance (F1 score) under progressive multi-turn semantic edits.}
    \label{fig:abla_multi_edits}
\end{figure}

This experiment evaluates CapsFake under a progressive attack scenario, where semantic manipulations are applied over multiple rounds, simulating persistent or evolving adversarial behaviors. We leverage a subset of the SEED-Data-Edit dataset~\cite{geSEEDDataEditTechnicalReport2024}, which contains images edited over 1 to 5 rounds, with each round introducing increasingly complex and diverse modifications. This setting mirrors realistic threat environments in which malicious actors iteratively refine image alterations to bypass detection systems.

Figure~\ref{fig:abla_multi_edits} reports the F1 scores of CapsFake and five baseline methods across varying edit rounds. CapsFake demonstrates clear and consistent superiority across all edit rounds, achieving F1 scores of 73.48\%, 87.00\%, 90.92\%, 95.72\%, and 96.37\% from 1 to 5 edits, respectively. This progression highlights the model’s ability to capture both subtle and accumulative semantic inconsistencies through its multimodal capsule routing architecture. Compared to the strongest baseline, UnivConv2B, which achieves 68.60\% at one edit and peaks at 90.00\% after five edits, CapsFake offers a 4.88\% margin at the lowest manipulation level and continues to outperform in later stages. Methods like DE-FAKE and ObjectFormer improve significantly under heavier manipulations (e.g., DE-FAKE rising from 48.01\% to 82.48\%), but remain consistently behind CapsFake. In contrast, DCT shows limited adaptability, plateauing around 59\%, even under multiple edits. 

This robustness to multi-stage attacks is critical for secure deployment in real-world media forensics pipelines, where edits may evolve across time or through collaborative adversarial refinement. Further analysis and ablation results are provided in Appendix~\ref{sec:full_multi_turn_edit}.

\section{Generalization Evaluation}
Robust generalization to unseen domains is a critical requirement for secure deployment of deepfake detectors in real-world environments, where manipulation styles and data sources are continuously evolving. In this experiment, we evaluate the cross-domain transferability of CapsFake and baseline models under domain shift by training on the \textit{One-turn Edits} dataset (denoted \textit{One}) and testing on two unseen datasets: \textit{MagicBrush} (\textit{Magic}) and \textit{Open Images Edits} (\textit{Open}).

The \textit{One} dataset consists of single-edit manipulations sourced from Unsplash~\cite{zahidalichesserlukeandcarbonetimothyUnsplash2023}, SAM~\cite{kirillovSegmentAnything2023}, and JourneyDB~\cite{sunJourneyDBBenchmarkGenerative2023}, and serves as a training set with subtle and localized edits that challenge detection models. We evaluate models under three scenarios: (i) \textit{One} $\rightarrow$ \textit{One} (in-domain generalization), (ii) \textit{One} $\rightarrow$ \textit{Magic}, and (iii) \textit{One} $\rightarrow$ \textit{Open} (cross-domain transfer).

\begin{table}
\centering
\caption{Transferability performance (F1 score) on unseen image editing data.}\label{tab:transferability}
\resizebox{\columnwidth}{!}{%
\begin{tabular}{cccc}
\toprule 
Methods & One $\rightarrow$ One & One $\rightarrow$ Magic & One $\rightarrow$ Open \tabularnewline
\midrule 
UniCLIP & 52.07 & 59.14 & 60.02\tabularnewline
DCT & 59.16 & 53.68 & 59.98 \tabularnewline
DE-FAKE & 48.01 & 53.03 & 50.19\tabularnewline
ObjectFormer & 54.21 & 57.26 & 49.41\tabularnewline
UnivConv2B & 68.60 & 54.16 & 53.77\tabularnewline
\midrule 
\textbf{CapsFake} & \textbf{73.48} & \textbf{70.38} & \textbf{69.36}\tabularnewline
\bottomrule 
\end{tabular}
}%
\end{table}

As shown in Table~\ref{tab:transferability}, CapsFake consistently achieves the highest F1 scores across all three settings. For in-domain evaluation (\textit{One} $\rightarrow$ \textit{One}), CapsFake reaches 73.48\%, outperforming the strongest baseline UnivConv2B (68.60\%). Under distribution shift, CapsFake’s advantage becomes more pronounced: on \textit{MagicBrush}, it scores 70.38\%, exceeding all baselines by at least 11.24\% compared to the next best method (UnivCLIP at 59.14\%). On \textit{Open Images Edits}, CapsFake again leads with 69.36\%, marking a 9.34\% improvement over UnivCLIP (60.02\%).

These results highlight CapsFake’s robustness to both intra-domain complexity and inter-domain variation, demonstrating its resilience to unseen manipulation styles and content distributions. In contrast, baseline models such as DE-FAKE and ObjectFormer exhibit notable performance drops in cross-domain scenarios, signaling limited adaptability.

\begin{figure*}
\begin{centering}
\resizebox{0.6\textwidth}{!}{%
\includegraphics{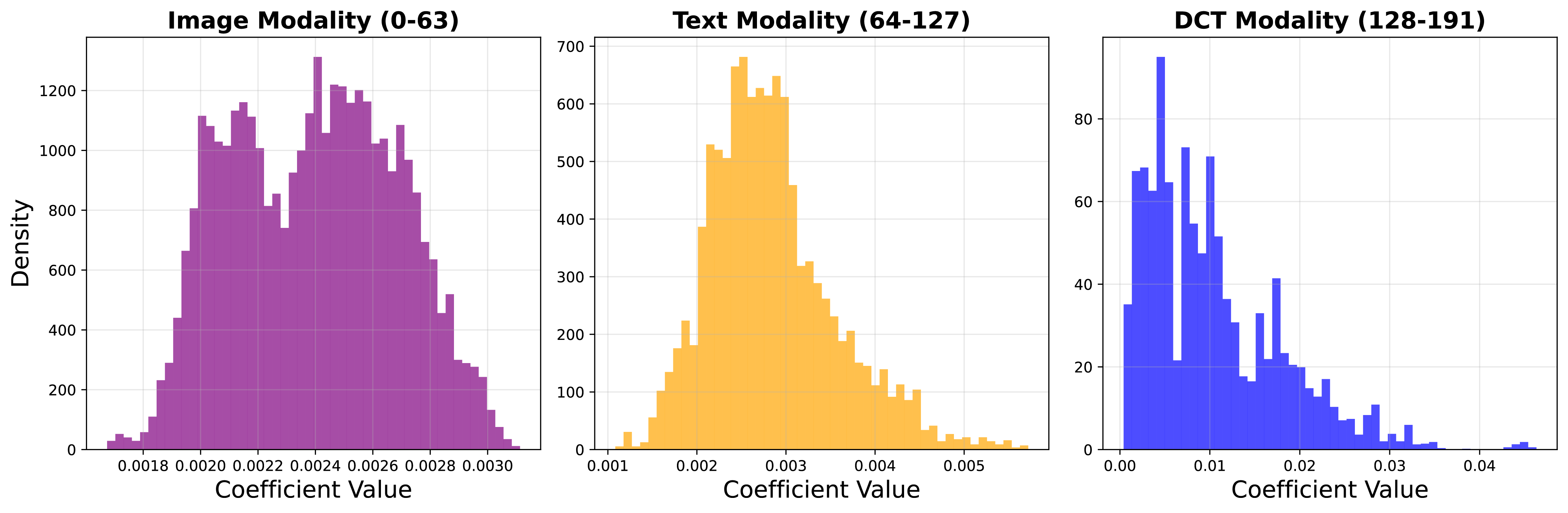}
}
\caption{Visualization of coupling coefficients $\alpha_{i,k}$ across three modalities from the MagicBrush dataset.}
\label{fig:abla_routing_histogram}
\par\end{centering}
\end{figure*}

\begin{figure*}
\begin{centering}
\resizebox{0.9\textwidth}{!}{%
\includegraphics{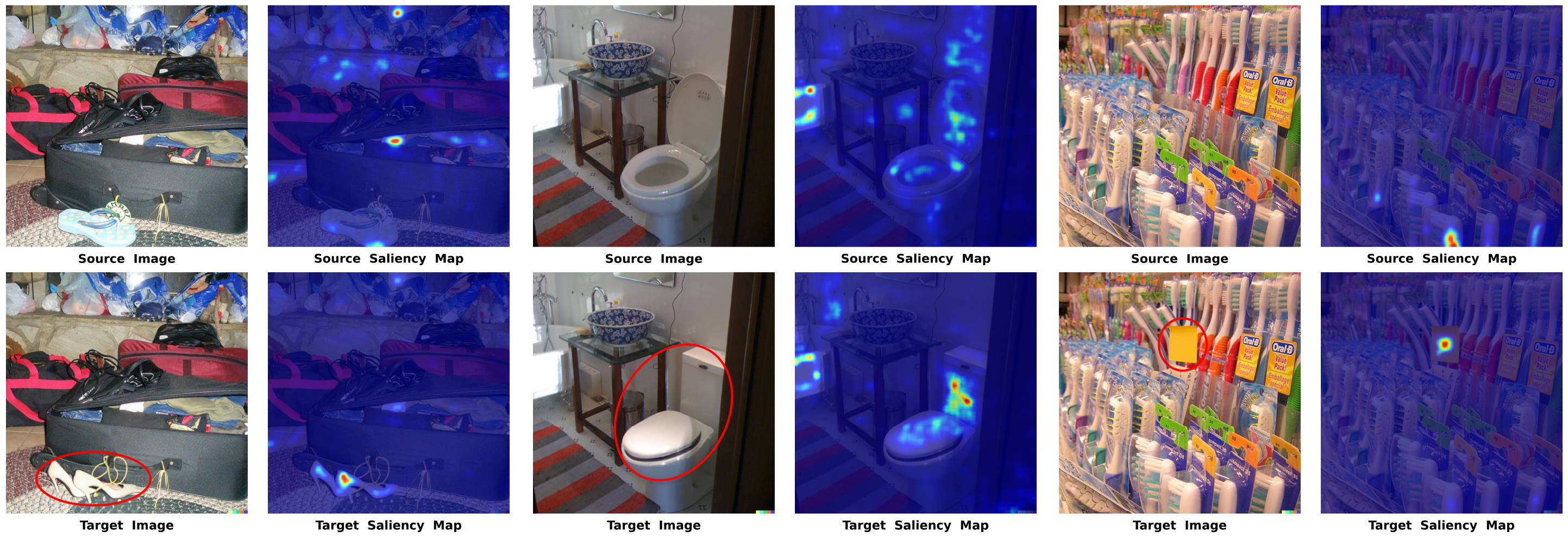}
}
\caption{Visualization of saliency maps for real (source) and manipulated (target) images from the MagicBrush dataset, with red circles highlighting the manipulated regions.}
\label{fig:abla_saliency}
\par\end{centering}
\end{figure*}

\section{Modality Analysis}
\begin{table}
\centering
\caption{Deepfake detection performance (\%) for different modality combinations on the Open Images Edits.}
\label{tab:modality-ablation}
\begin{tabular}{lc}
\hline
Modality Combination & Recall (\%) \\
\hline
Visual + Text        & 94.79 \\
Visual + Frequency         & 95.55 \\
Text + Frequency          & 84.36 \\
\textbf{Visual + Text + Frequency} & \textbf{95.74} \\
\hline
\end{tabular}
\end{table}
\subsection{Modality Importance}
Table~\ref{tab:modality-ablation} presents a modality ablation study assessing the contribution of each input signal: visual, textual, and frequency, to overall detection performance on the Open Images Edits dataset. This experiment evaluates the robustness and complementary nature of each modality in identifying semantically manipulated content.

Among the bi-modal configurations, the combination of \textit{Visual + Frequency} achieves the highest recall (95.55\%), closely followed by \textit{Visual + Text} (94.79\%). These results suggest that visual features are the dominant signal for deepfake detection, while frequency-based cues (e.g., DCT artifacts) provide critical complementary information, especially for detecting low-level editing inconsistencies such as compression traces or unnatural textures.

In contrast, the \textit{Text + Frequency} configuration results in a substantial drop in performance (84.36\%), indicating that in the absence of spatial visual context, the remaining modalities are insufficient for reliable classification. This underscores the importance of visual grounding when interpreting textual prompts or frequency-domain anomalies.

The full tri-modal configuration (\textit{Visual + Text + Frequency}) achieves the highest recall at 95.74\%, demonstrating the benefit of multimodal fusion. These findings underscore the importance of architectural diversity in modality integration. Relying on a single or reduced modality set introduces potential blind spots that adversaries may exploit. CapsFake’s ability to effectively leverage and fuse heterogeneous signals enhances its robustness against attacks targeting specific input pathways.

\begin{figure*}
\begin{centering}
\resizebox{0.8\textwidth}{!}{%
\includegraphics{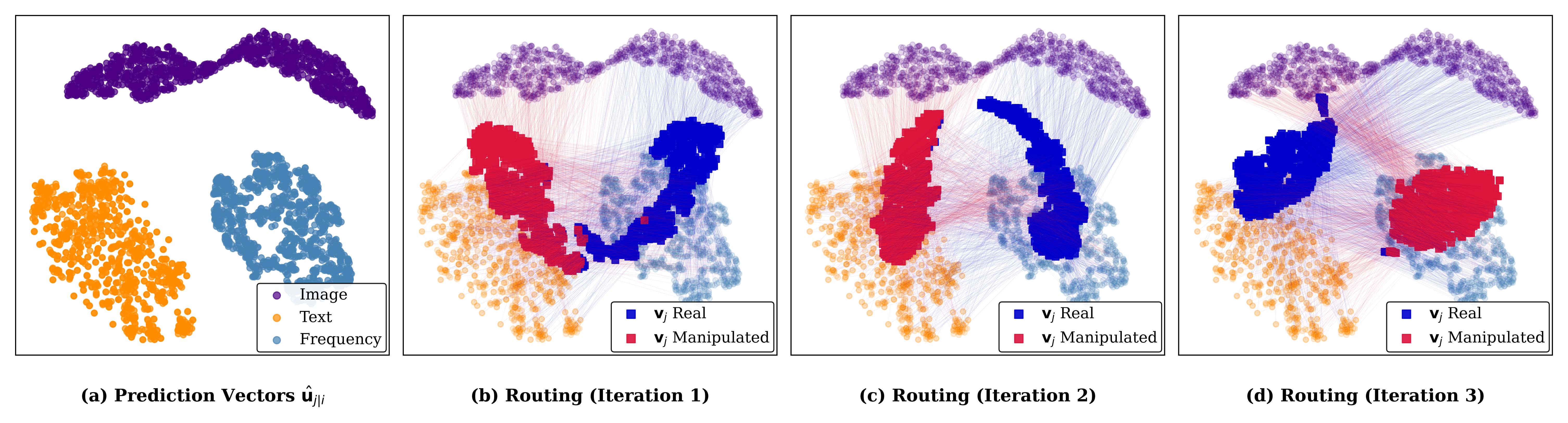}
}
\caption{\textit{t}-SNE visualization of capsules before and after routing within the DetectCaps layer using the MagicBrush dataset.}
\label{fig:abla_tsne}
\par\end{centering}
\end{figure*}

\subsection{Coupling Coefficients Visualization}
The coupling coefficients $\alpha_{i,k}$ are central to the dynamic routing process in the DetectCaps layer, determining the strength of the connection between lower-layer and higher-layer capsules. Visualizing their distribution offers insights into how CapsFake routes and integrates modality-specific information for deepfake detection.

Figure~\ref{fig:abla_routing_histogram} shows a histogram of coupling coefficients computed from all test images in the MagicBrush dataset. Each prediction produces $3 \times 64 = 192$ coefficients corresponding to capsules from three modalities: visual, text, and frequency. The visual modality exhibits the most concentrated coefficient range (0.0018–0.0030) and the highest peak density (greater than 1200), indicating strong and consistent contributions of spatial and contextual visual features, critical for detecting object-level manipulations. The text modality spans a slightly wider range (0.001–0.005) with a moderate peak around 700, reflecting its role in modeling semantic intent from editing prompts. In contrast, the frequency modality shows the widest spread (0.0–0.04) and a low peak density (around 90), suggesting sparse but important responses to localized frequency artifacts, such as compression anomalies or texture inconsistencies.

These results suggest that while visual features offer a reliable structural foundation, the text and frequency modalities provide complementary and specialized cues. CapsFake’s routing process effectively balances these inputs, enhancing robustness across diverse manipulation types.

\section{Saliency and Localization}
\label{sec:saliency_maps}

Saliency maps~\cite{simonyanDeepConvolutionalNetworks2014} provide visual explanations for model predictions by identifying the regions of an image that most influence classification outcomes. In security-critical applications like deepfake detection, such interpretability is essential for validating and auditing automated decisions. We use saliency analysis to evaluate whether CapsFake accurately identifies manipulated regions across diverse semantic edits.

To compute saliency, we define the prediction confidence as \( p = \max \left\{ \left\| \mathbf{v}_r \right\|, \left\| \mathbf{v}_f \right\| \right\} \), where \( \mathbf{v}_r \) and \( \mathbf{v}_f \) denote the real and manipulated class capsules from the DetectCaps layer. The gradient of \( p \) w.r.t. the input image \( \mathbf{X} \in \mathbb{R}^{H \times W \times C} \) is:
\[
\mathbf{G} = \left| \frac{\partial p}{\partial \mathbf{X}} \right|,
\]
and the saliency at each pixel is computed as:
\[
S(x, y) = \frac{1}{C} \sum_{c=1}^{C} \mathbf{G}(x, y, c).
\]
Figure~\ref{fig:abla_saliency} shows saliency visualizations across three representative samples. In each case, the source (real) image and corresponding saliency map are shown in the top row, while the manipulated (target) image and saliency are in the bottom row. In the first column, a pair of shoes is synthetically inserted into the scene. The saliency map for the manipulated image sharply highlights the inserted shoes, while the real image’s saliency is diffused, focusing on the overall scene context. Similarly, in the second column, a toilet lid has been added. The manipulated saliency map shows strong attention on this new object, while the real saliency focuses on the existing bathroom fixtures. In the third example, a new price label is added among toothbrushes, CapsFake sharply localizes the added tag, even within a cluttered background.

These results confirm that CapsFake effectively identifies and localizes subtle, semantically relevant manipulations. The consistency of the saliency focus on manipulated regions (across object insertions and contextual changes) demonstrates not only high detection accuracy, but also model transparency. This interpretability is essential for security contexts, where human analysts may rely on the model’s focus regions for forensic verification or evidence review. The ability to attribute predictions to specific localized content enhances the trustworthiness of CapsFake under adversarial conditions.



\section{Routing Process Visualization}

To evaluate how dynamic routing contributes to CapsFake’s decision-making and robustness, we visualize the evolution of capsule interactions during inference using \textit{t}-SNE~\cite{JMLR:v9:vandermaaten08a}. Specifically, we project the prediction vectors in $\hat{\mathbf{U}}$ and the final class capsules $\mathbf{v}_k$ produced by the DetectCaps layer across multiple routing iterations.

Figure~\ref{fig:abla_tsne} depicts this progression using test images from the MagicBrush dataset. Subfigure~\ref{fig:abla_tsne}a color-codes the prediction vectors by modality: visual (purple), text (orange), and frequency (cyan), revealing an initial mixture of features across modalities. Subfigures~\ref{fig:abla_tsne}b–\ref{fig:abla_tsne}d then show the evolution of the final class capsules over three routing iterations. After the first iteration, real (blue) and manipulated (red) class capsules remain largely overlapping, indicating weak semantic separation. By the second and third iterations, routing progressively refines the signal alignment, producing a distinct separation between real and manipulated examples.

This iterative refinement highlights CapsFake’s ability to selectively amplify modality-specific features that are semantically consistent with the target class. The dynamic routing mechanism not only enforces feature relevance but also facilitates cross-modal disentanglement, crucial in detecting subtle, context-aware manipulations. This property is especially valuable in adversarial scenarios, where edits may be spatially sparse or semantically ambiguous.

\section{Conclusion}

Instruction-guided image editing introduces a critical threat to digital media integrity by enabling context-aware manipulations that are both semantically precise and visually coherent. These edits often evade human perception and existing detection systems, posing serious risks in domains such as misinformation, identity fraud, and evidentiary tampering. In this paper, we introduced \textbf{CapsFake}, a novel multimodal capsule network designed to detect deepfake image edits with high precision and robustness. CapsFake integrates low-level capsules from visual, textual, and frequency modalities and employs a dynamic routing mechanism to construct high-level semantic capsules that specialize in identifying subtle manipulations. We evaluated CapsFake across multiple benchmarks, including MagicBrush, Unsplash Edits, Open Images Edits, and multi-turn SEED edits, and observed consistent performance improvements over state-of-the-art baselines, with up to 20\% gains in accuracy. Our model demonstrated strong resilience to natural image perturbations, adversarial attacks (white-box and black-box), and unseen domain shifts. Notably, CapsFake retained near-perfect recall under PGD adversarial settings and exhibited strong generalization on cross-domain datasets and progressive multi-turn editing scenarios. Beyond empirical performance, we conducted visualization-based analyses, including saliency maps and coupling coefficient distributions, to provide insight into the model’s decision-making process. These findings support CapsFake’s interpretability and its suitability for use in forensic and security-critical contexts. In summary, our work presents an effective, interpretable, and robust solution for semantic deepfake detection, bridging the gap between visual fidelity and manipulation intent. CapsFake offers a promising defense architecture for safeguarding visual media integrity in the face of increasingly sophisticated generative editing techniques.

\bibliographystyle{IEEEtran}
\bibliography{References_v2}

\newpage

\appendix

\begin{table*}
\centering{}\caption{Detailed dataset information used in our main experiments and ablation
studies.}\label{tab:dataset-details}
\begin{tabular}{ccccccc}
\hline
Datasets & Type & Train & Validation & Test & Total Images & Image size\tabularnewline
\hline
\multirow{2}{*}{MagicBrush} & Source & 4,512 & 266 & 535 & 5,313 & 500$\times$500\tabularnewline
\cline{2-7}
 & Target & 4,512 & 266 & 535 & 5,313 & 1024$\times$1024\tabularnewline
\hline
\multirow{2}{*}{Unsplash Edits} & Source & 12,000 & 1,500 & 1,500 & 15,000 & 512$\times$512\tabularnewline
\cline{2-7}
 & Target & 12,000 & 1,500 & 1,500 & 15,000 & 512$\times$512\tabularnewline
\hline
\multirow{2}{*}{Open Images Edits} & Source & 20,857 & 2,607 & 2,608 & 26,072 & 768$\times$768\tabularnewline
\cline{2-7}
 & Target & 20,857 & 2,607 & 2,608 & 26,072 & 768$\times$768\tabularnewline
\hline
\multirow{2}{*}{Multi-turn (1 Edit)} & Source & 14,538 & 1,817 & 1,811 & 18,166 & varies\tabularnewline
\cline{2-7}
 & Target & 14,538 & 1,817 & 1,811 & 18,166 & varies\tabularnewline
\hline
\multirow{2}{*}{Multi-turn (2 Edits)} & Source & 14,844 & 1,855 & 1,856 & 18,555 & varies\tabularnewline
\cline{2-7}
 & Target & 14,844 & 1,855 & 1,856 & 18,555 & varies\tabularnewline
\hline
\multirow{2}{*}{Multi-turn (3 Edits)} & Source & 14,538 & 1,817 & 1,818 & 18,173 & varies\tabularnewline
\cline{2-7}
 & Target & 14,538 & 1,817 & 1,818 & 18,173 & varies\tabularnewline
\hline
\multirow{2}{*}{Multi-turn (4 Edits)} & Source & 9,248 & 1,156 & 1,156 & 11,560 & varies\tabularnewline
\cline{2-7}
 & Target & 9,248 & 1,156 & 1,156 & 11,560 & varies\tabularnewline
\hline
\multirow{2}{*}{Multi-turn (5 Edits)} & Source & 3,223 & 402 & 404 & 4,029 & varies\tabularnewline
\cline{2-7}
 & Target & 3,223 & 402 & 404 & 4,029 & varies\tabularnewline
\hline
\end{tabular}
\end{table*}

\section{Dataset information}\label{sec:dataset_information}
Table \ref{tab:dataset-details} provides detailed information about the datasets used in our main experiments and ablation studies. The datasets include MagicBrush, Unsplash Edits, Open Images Edits, and Multi-turn datasets, each featuring varying numbers of edits. The source (original) and target (manipulated) images in each dataset are divided into training, validation, and test sets. Additionally, the total number of images and their sizes are detailed for each dataset, with the combined total across all datasets amounting to 233,736 images.

\section{Additional Results and Discussion}\label{sec:additional_results}

\subsection{Full results on Multi-turn Edits}\label{sec:full_multi_turn_edit}
The results in Figure \ref{fig:abla_multi_edits} illustrate the detection performance evaluated using the F1 score. Additional metrics, including precision, recall, and accuracy for each dataset, are provided in Tables \ref{tab:1-turn}, \ref{tab:2-turn}, \ref{tab:3-turn}, \ref{tab:4-turn}, and \ref{tab:5-turn}. Details of the training, validation, and test splits for the multi-turn editing datasets are outlined in Table \ref{tab:dataset-details}.

The results consistently demonstrate that CapsFake outperforms baseline methods across all datasets. In Table \ref{tab:1-turn}, CapsFake achieves the highest F1 score of 73.48\%, surpassing the closest baseline (UnivConv2B at 68.6\%) by a margin of 4.88\%. In Table \ref{tab:2-turn}, CapsFake excels with an F1 score of 87.00\%, significantly outperforming UnivConv2B (74.19\%) with an improvement of 12.81\%. Table \ref{tab:3-turn} shows that CapsFake achieves an F1 score of 90.92\%, a 10.85\% improvement over UnivConv2B (80.07\%), with precision and recall also exceeding 90\%. In Tables \ref{tab:4-turn} and \ref{tab:5-turn}, CapsFake records the highest F1 scores of 95.72\% and 96.37\%, outperforming UnivConv2B by margins of 8.38\% and 6.37\%, respectively. These results highlight the robustness of CapsFake in detecting deepfake image edits across diverse scenarios, such as removing, adding, or replacing objects; changing actions; altering text or patterns; and modifying object counts. Its superior performance across all datasets underscores its ability to effectively leverage multimodal information and dynamically adapt to complex and varied editing scenarios.

\subsection{Results on different design of CapsFake}\label{sec:design_capsfake}
Table \ref{tab:capsfake-design} presents the performance of different CapsFake designs. First, we evaluate training CapsFake from \textbf{S}cratch (CapsFake-S). This architecture processes a 320 $\times$ 320 input image through a convolutional backbone followed by a Primary Capsule layer \cite{sabourDynamicRoutingCapsules2017} to extract hierarchical and spatially rich feature representations. The convolutional backbone consists of three layers, progressively reducing spatial dimensions from 320 $\times$ 320 to 40 $\times$ 40 while increasing the channel depth to 256. These features are then passed to the Primary Capsule layer, which follows the same architecture as the original Capsule Network \cite{sabourDynamicRoutingCapsules2017}. In this design, CapsFake-S achieves a precision of 67.33\%, recall of 47.01\%, F1 score of 55.36\%, and accuracy of 62.10\%. Next, we examine training CapsFake using a pretrained \textbf{F}eature volume (CapsFake-F). In this design, the convolutional backbone is the pretrained OpenCLIP-ConvNextLarge \cite{LaionCLIPconvnext_large_d_320laion2Bs29Bb131KftsoupHugging2023}, which is frozen during training. In this design, the convolutional backbone is the pretrained OpenCLIP-ConvNextLarge \cite{LaionCLIPconvnext_large_d_320laion2Bs29Bb131KftsoupHugging2023}, which remains frozen during training. The feature volume, with a shape of \((1536, 10, 10)\) from the last convolutional layer of the backbone, is passed to the Primary Capsule layer. CapsFake-F achieves a precision of 72.35\%, recall of 61.62\%, F1 score of 66.56\%, and accuracy of 69.06\%. Finally, we evaluate CapsFake with a pretrained embedding \textbf{V}ector (CapsFake-V), which is also the current design of CapsFake. This architecture achieves the best performance, with a precision of 89.19\%, recall of 80.98\%, F1 score of 84.89\%, and accuracy of 85.58\%. These results demonstrate that leveraging pretrained embedding vectors significantly improves the detection performance of CapsFake. While CapsFake-S is trained from scratch and consists of only three convolutional layers, this design may not be sufficient to capture complex features or subtle manipulations in the images. CapsFake-F, which uses a pretrained backbone to extract feature volumes, shows better performance but is prone to overfitting in complex natural images containing multiple objects and intricate backgrounds. This occurs because low-level capsules focus on background patterns and local object features, which hinders their ability to generalize and detect subtle manipulations. CapsFake-V, on the other hand, leverages pretrained embedding vectors that capture the most relevant and discriminative features, effectively avoiding noisy background textures and focusing on distinguishing features related to manipulated areas. Thus, CapsFake-V is chosen as the architecture for the proposed CapsFake due to its superior performance and robustness.

\begin{table}
\centering{}\caption{Deepfake detection performance on Open Images Edits across different
architectural designs.}\label{tab:capsfake-design}
\begin{tabular}{ccccc}
\hline
Methods & Precision & Recall & F1 & Accuracy\tabularnewline
\hline
CapsFake-S & 67.33 & 47.01 & 55.36 & 62.10\tabularnewline
CapsFake-F & 72.35 & 61.62 & 66.56 & 69.06\tabularnewline
CapsFake-V & \textbf{89.19} & \textbf{80.98} & \textbf{84.89} & \textbf{85.58}\tabularnewline
\hline
\end{tabular}
\end{table}

\begin{table}
\centering{}\caption{Deepfake detection performance on One-turn Edits.}\label{tab:1-turn}
\begin{tabular}{ccccc}
\hline
Methods & Precision & Recall & F1 & Accuracy\tabularnewline
\hline
UnivCLIP & 60.96 & 45.44 & 52.07 & 58.17\tabularnewline
DCT & 51.65 & 69.24 & 59.16 & 52.21\tabularnewline
DE-FAKE & 60.85 & 39.65 & 48.01 & 57.07\tabularnewline
ObjectFormer & 54.29 & 54.14 & 54.21 & 54.28\tabularnewline
UnivConv2B & 61.04 & 78.3 & 68.60 & 64.16\tabularnewline
\hline
\textbf{CapsFake} & \textbf{69.80} & \textbf{77.58} & \textbf{73.48} & \textbf{72.00}\tabularnewline
\hline
\end{tabular}
\end{table}

\begin{table}
\centering{}\caption{Deepfake detection performance on Two-turn Edits.}\label{tab:2-turn}
\begin{tabular}{ccccc}
\hline
Methods & Precision & Recall & F1 & Accuracy\tabularnewline
\hline
UnivCLIP & 67.22 & 56.39 & 61.33 & 64.46\tabularnewline
DCT & 53.01 & 69.33 & 60.08 & 53.95\tabularnewline
DE-FAKE & 73.76 & 47.28 & 57.62 & 65.24\tabularnewline
ObjectFormer & 64.79 & 58.16 & 61.29 & 63.27\tabularnewline
UnivConv2B & 66.61 & 83.73 & 74.19 & 70.87\tabularnewline
\hline
\textbf{CapsFake} & \textbf{87.81} & \textbf{86.20} & \textbf{87.00} & \textbf{87.12}\tabularnewline
\hline
\end{tabular}
\end{table}
\begin{table}
\centering{}\caption{Deepfake detection performance on Three-turn Edits.}\label{tab:3-turn}
\begin{tabular}{ccccc}
\hline
Methods & Precision & Recall & F1 & Accuracy\tabularnewline
\hline
UnivCLIP & 77.11 & 65.40 & 70.77 & 72.99\tabularnewline
DCT & 57.23 & 62.93 & 59.94 & 57.95\tabularnewline
DE-FAKE & 81.01 & 62.16 & 70.34 & 73.79\tabularnewline
ObjectFormer & 75.37 & 63.81 & 69.11 & 71.48\tabularnewline
UnivConv2B & 75.10 & 85.75 & 80.07 & 78.66\tabularnewline
\hline
\textbf{CapsFake} & \textbf{95.38} & \textbf{90.92} & \textbf{90.92} & \textbf{93.26}\tabularnewline
\hline
\end{tabular}
\end{table}
\begin{table}
\centering{}\caption{Deepfake detection performance on Four-turn Edits.}\label{tab:4-turn}
\begin{tabular}{ccccc}
\hline
Methods & Precision & Recall & F1 & Accuracy\tabularnewline
\hline
UnivCLIP & 87.16 & 78.72 & 82.73 & 83.56\tabularnewline
DCT & 58.66 & 60.03 & 59.34 & 58.87\tabularnewline
DE-FAKE & 90.69 & 72.49 & 80.58 & 82.53\tabularnewline
ObjectFormer & 87.54 & 77.16 & 82.02 & 83.10\tabularnewline
UnivConv2B & 84.72 & 90.14 & 87.34 & 86.94\tabularnewline
\hline
\textbf{CapsFake} & \textbf{98.80} & \textbf{92.82} & \textbf{95.72} & \textbf{95.85}\tabularnewline
\hline
\end{tabular}
\end{table}
\begin{table}
\centering{}\caption{Deepfake detection performance on Five-turn Edits.}\label{tab:5-turn}
\begin{tabular}{ccccc}
\hline
Methods & Precision & Recall & F1 & Accuracy\tabularnewline
\hline
UnivCLIP & 87.13 & 80.45 & 83.66 & 84.28\tabularnewline
DCT & 60.28 & 52.97 & 56.39 & 59.03\tabularnewline
DE-FAKE & 90.53 & 75.74 & 82.48 & 83.91\tabularnewline
ObjectFormer & 87.58 & 71.36 & 78.64 & 80.62\tabularnewline
UnivConv2B & 88.70 & 91.34 & 90.00 & 89.85\tabularnewline
\hline
\textbf{CapsFake} & \textbf{97.47} & \textbf{95.3} & \textbf{96.37} & \textbf{96.41}\tabularnewline
\hline
\end{tabular}
\end{table}

\begin{figure*}
\begin{centering}
\resizebox{0.8\textwidth}{!}{%
\includegraphics{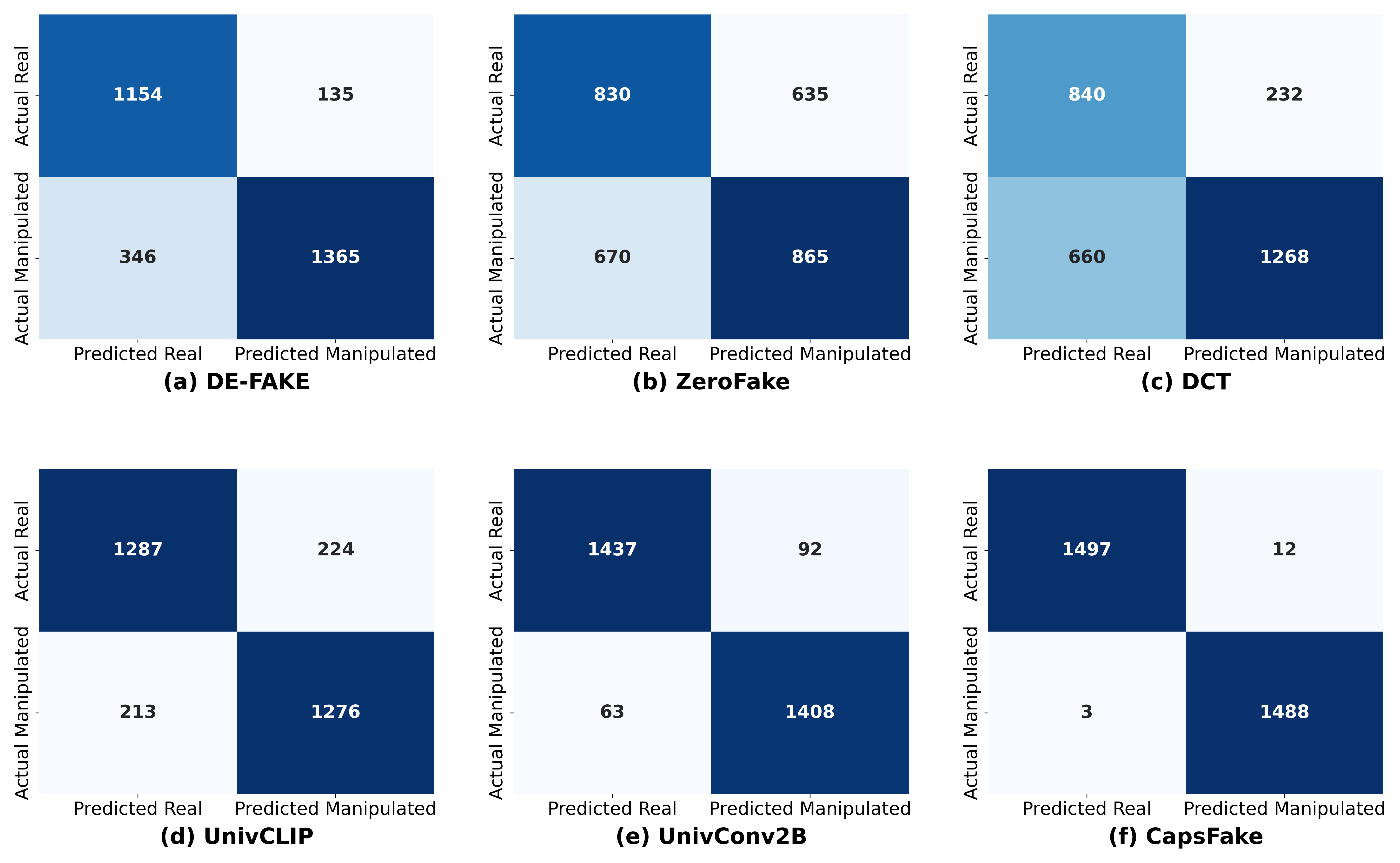}
}
\caption{Confusion matrices of detection models evaluated on 3,000 test images from the Unsplash dataset.}
\label{fig:confusion_matrix}
\par\end{centering}
\end{figure*}

\subsection{Confusion matrix analysis}\label{sec:confusion_matrix}
Figure \ref{fig:confusion_matrix} presents the confusion matrices for detectors evaluated on 3,000 test images from the Unsplash dataset. Overall, models such as DE-FAKE, ZeroFake, DCT, and UnivCLIP demonstrate poor performance in detecting deepfake images, with a high number of false positives (over 130) and false negatives (over 200). In contrast, UnivConv2B and CapsFake show significantly better performance, with notably lower false positives and false negatives. UnivConv2B achieves strong results, with high true positive (1,408) and true negative (1,437) counts, though it still records 92 false positives and 63 false negatives. CapsFake, however, demonstrates the best performance, achieving an excellent balance of true positives (1,488) and true negatives (1,497), while minimizing false positives (12) and false negatives (3). These results highlight the robustness of CapsFake, showcasing exceptional precision and recall in detecting even subtle manipulations.


\end{document}